%% file: acl_latex.tex
\documentclass[11pt]{article}

\usepackage[final]{acl}

\usepackage{tgtermes}
\usepackage{latexsym}

\usepackage[T1,T5]{fontenc} %

\usepackage[utf8]{inputenc}

\usepackage{microtype}

\usepackage{inconsolata}

\usepackage{graphicx}

\usepackage[cjk]{kotex}
\usepackage[normalem]{ulem}
\usepackage{amsmath}
\usepackage{amssymb}
\usepackage{array}
\usepackage{bm}
\usepackage{booktabs}
\usepackage{CJKutf8}
\usepackage{colortbl}
\usepackage{diagbox}
\usepackage{footmisc}
\usepackage{lipsum}
\usepackage{makecell}
\usepackage{mdframed}
\usepackage{multirow}
\usepackage{pifont}
\usepackage{soul}
\usepackage{subcaption}
\usepackage{tabularx}
\usepackage{tikz}
\usepackage{upquote}
\usepackage{arydshln}
\usepackage[most]{tcolorbox}
\usepackage{menukeys}
\usepackage{cleveref}
\usepackage{xspace}
\usepackage{rotating}

\graphicspath{ {./images/} }
\mdfsetup{
  linecolor=white,
  backgroundcolor=gray!20,
}
\newcolumntype{?}{!{\vrule width 1.2pt}}
\useunder{\uline}{\ul}{}
\crefformat{section}{\S#2#1#3}
\crefformat{subsection}{\S#2#1#3}
\crefformat{subsubsection}{\S#2#1#3}

\def\eg{\emph{e.g}.,\xspace}
\def\ie{\emph{i.e}.,\xspace}

\newcommand{\cb}[2]{\colorbox[HTML]{#1}{#2}}
\definecolor{lightblue}{HTML}{C5E0ED}
\definecolor{lightyellow}{HTML}{F3D7A3}
\definecolor{lightgray}{HTML}{E0E0E0}

\def\docommandbetter#1 {\colorbox{lightblue!100}{#1} \let\next\argii}
\def\argii{\let\next\docommandbetter}

\def\docommandworse#1 {\colorbox{lightyellow!100}{#1} \let\next\argii}
\def\argii{\let\next\docommandworse}
\newcommand\notsig{\def\argii{}\docommandnotsig}
\def\docommandnotsig#1 {\colorbox{lightgray!100}{#1} \let\next\argii}
\def\argii{\let\next\docommandnotsig}

\newcommand{\chuhan}{{\fontencoding{T5}\selectfont{Chữ Hán}}}

\newcommand{\cjk}[1]{
  \begin{CJK*}{UTF8}{bsmi}{#1}
\end{CJK*}}
\newcommand{\cjkj}[1]{
  \begin{CJK*}{UTF8}{min}{#1}
\end{CJK*}}

\newcommand{\ztexttt}[1]{%
  {\fontencoding{T1}\fontfamily{zi4}\selectfont#1}%
}
\newcommand{\btexttt}[1]{\textcolor{darkblue}{\texttt{#1}}}

\title{Shared Heritage, Distinct Writing: \\Rethinking Resource Selection for East Asian Historical Documents}

\author{
  Seyoung Song$^\diamondsuit$
  \hspace{4mm}
  Haneul Yoo$^\diamondsuit$
  \hspace{4mm}
  Jiho Jin$^\diamondsuit$
  \hspace{4mm}
  Kyunghyun Cho$^{\spadesuit\clubsuit}$
  \hspace{4mm}
  Alice Oh$^\diamondsuit$
  \\
  \\
  $^{\diamondsuit}$KAIST
  \hspace{4mm}
  $^{\spadesuit}$New York University
  \hspace{4mm}
  $^{\clubsuit}$Genentech
  \\
  \\
  \ztexttt{\{\href{mailto:seyoung.song@kaist.ac.kr}{\color{black}{seyoung.song}}, \href{mailto:haneul.yoo@kaist.ac.kr}{\color{black}{haneul.yoo}}, \href{mailto:jinjh0123@kaist.ac.kr}{\color{black}{jinjh0123}}\}@kaist.ac.kr,}\\
  \ztexttt{kyunghyun.cho@nyu.edu,}
  \ztexttt{alice.oh@kaist.edu}
}

\begin{document}
\maketitle
\begin{abstract}
  \input{contents/0_abstract}
\end{abstract}

\section{Introduction}
\input{contents/1_introduction}

\section{Background}
\input{contents/2_background}

\section{Experiments}
\input{contents/3_experiments}

\section{Discussions}
\input{contents/4_discussion}

\section{Related Work}
\input{contents/5_related_work}

\section{Conclusion}

\input{contents/6_conclusion}

\section*{Limitations}

\input{contents/7_limitation}

\section*{Ethical Considerations}
\input{contents/8_ethics}

\section*{Acknowledgments}
\input{contents/9_ack}


\input{acl_latex.bbl}
\clearpage
\appendix

\section*{Appendix}

\input{contents/appendix}

\end{document}

%% file: contents/0_abstract.tex
Historical documents in the Sinosphere are known to share common formats and practices, particularly in veritable records compiled by court historians.
This shared linguistic heritage has led researchers to use Classical Chinese resources for cross-lingual transfer when processing historical documents from Korea and Japan, which remain relatively low-resource.
In this paper, we question the assumption of cross-lingual transferability from Classical Chinese to Hanja and Kanbun, the ancient written languages of Korea and Japan, respectively.
Our experiments across machine translation, named entity recognition, and punctuation restoration tasks show minimal impact of Classical Chinese datasets on language model performance for ancient Korean documents written in Hanja, with performance differences within $\pm{}0.0068$ F1-score for sequence labeling tasks and up to $+0.84$ BLEU score for translation.
These limitations persist consistently across various model sizes, architectures, and domain-specific datasets.
Our analysis reveals that the benefits of Classical Chinese resources diminish rapidly as local language data increases for Hanja, while showing substantial improvements only in extremely low-resource scenarios for both Korean and Japanese historical documents.
These findings emphasize the need for careful empirical validation rather than assuming benefits from indiscriminate cross-lingual transfer.

%% file: contents/1_introduction.tex
Classical Chinese served as a regional lingua franca across East Asia for over a millennium, where it was used to record government chronicles, literary works, and scientific discoveries.
These historical documents, particularly ``veritable records'' compiled by court historians, remain invaluable primary sources for studying the region's past.
As Classical Chinese spread throughout East Asia, it evolved into distinct writing systems---Hanja in Korea, Kanbun in Japan, and \chuhan{} in Vietnam---collectively forming the \emph{Sinosphere} or \emph{Chinese character cultural sphere}.

\input{figures/teaser_image}

\input{figures/mainrq}

Recent advances in natural language processing have enabled computational analysis of these historical documents, which is crucial as modern speakers can no longer directly interpret these ancient writings.
Researchers are increasingly leveraging Classical Chinese resources to develop language models for other Sinosphere languages~\cite[\xspace \emph{inter alia}]{yoo-etal-2022-hue, moon-2024-exploiting, wang-etal-2023-kanbun}.
This approach appears particularly promising given the shared literary traditions and significant resource disparity across these languages---with Classical Chinese being the most abundant, followed by Hanja, while Kanbun and \chuhan{} remain relatively scarce.
However, the effectiveness of such cross-lingual approaches has not been thoroughly evaluated, despite these writing systems having evolved independently over 1,500 years to accommodate distinct regional needs and cultural practices.

In this paper, we challenge this assumption by conducting comprehensive experiments across three tasks: machine translation (MT), named entity recognition (NER), and punctuation restoration (PR).
Figure~\ref{fig:mainrq} demonstrates that leveraging Classical Chinese corpora does not yield statistically significant improvements for NER and PR tasks across Hanja documents.
For MT, while there is a marginally positive effect (+0.84 BLEU score) for Hanja literary works, this improvement is not substantial---according to \citet{kocmi-etal-2024-navigating} and \citet{xu2024paradigm}, BLEU improvements of this magnitude typically correlate with human-perceived quality improvements in only 60-65\% of cases.
These results remain consistent across different model architectures and parameter scales, suggesting fundamental limitations in cross-lingual transfer between these historical languages (\cref{sec:model-scaling}).

To enable deeper analysis beyond the predominantly royal-centric Hanja research~\cite[\xspace \emph{inter alia}]{kang-etal-2021-restoring,yoo-etal-2022-hue,son-etal-2022-translating}, we introduce \emph{the Korean Literary Collections} (KLC), a corpus of literary works written in Hanja that captures diverse writing styles from individual scholars.
Our domain-specific analysis reveals that while incorporating Classical Chinese data shows mixed results overall, careful selection of similar writing styles---such as using Chinese classical poetry for Korean literary works---can lead to marginal improvements in translation performance (\cref{sec:domain-specific}).

Our investigation reveals that Classical Chinese resources provide benefit for only extremely low-resource scenarios, with their effectiveness diminishing rapidly as local language data increases for Hanja (\cref{sec:threshold-diminishing}).
Experiments with Japanese historical documents written in Kanbun show similar trends of effective cross-lingual transfer in low-resource settings (\cref{sec:mt-kanbun}).
Moreover, our vocabulary analyses across the Sinosphere show that character-level divergence is minimal, suggesting that the limited cross-lingual transferability stems from deeper linguistic differences (\cref{sec:vocab-divergence}).

Our findings across different dimensions emphasize that successful cross-lingual transfer in historical language processing requires considerations beyond shared writing systems, highlighting the importance of careful empirical validation that accounts for both resource availability and domain characteristics. Our contributions are as follows:
\begin{itemize}
  \item We question and empirically evaluate the efficacy of leveraging Classical Chinese resources for historical Asian language models.
  \item We demonstrate Classical Chinese integration yields minimal improvements for Hanja processing, while showing potential benefits for extremely low-resource scenarios.
  \item We provide analyses of cross-lingual transfer effectiveness that can inform the development of language models for historical documents across the Sinosphere.
  \item We publicly release our code and data, including the KLC dataset previously unexplored in the NLP community.\footnote{\ztexttt{\url{https://github.com/seyoungsong/Shared-Heritage-Distinct-Writing}}}
\end{itemize}

%% file: figures/teaser_image.tex
\begin{figure}[t]
  \centering
  \includegraphics[width=\columnwidth]{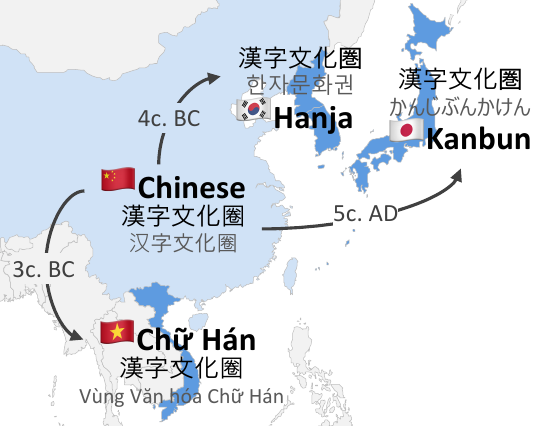}
  \caption{Language transfer from Classical Chinese to neighboring countries in Sinosphere. Classical Chinese had been transferred to neighboring countries in East Asia and used from the 6th century BC to the 20th century AD. While modern languages (\textit{gray}) are different from each other, ancient languages (\textit{black}) are mutually understandable.}
  \label{fig:teaser_image}
\end{figure}

%% file: figures/mainrq.tex
\begin{figure*}[t!]
  \centering
  \includegraphics[width=\textwidth]{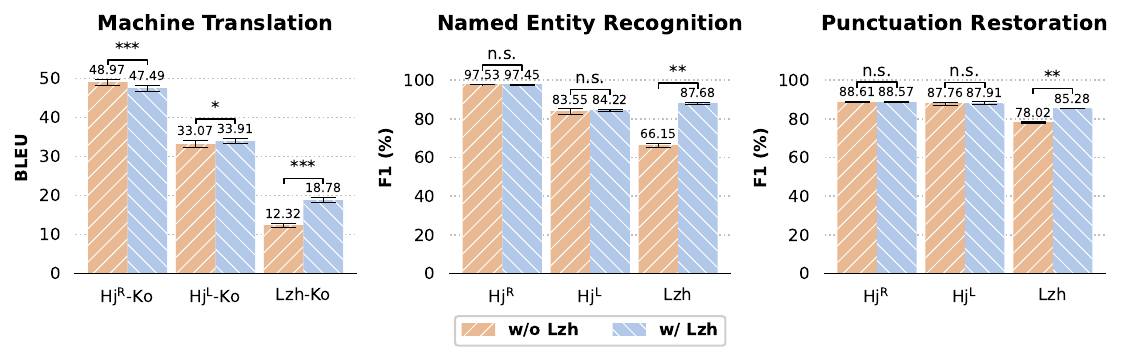}
  \caption{Comparison of models trained with and without Classical Chinese (Lzh). Results show BLEU scores (MT) and F1-scores (NER, PR) across three document types: Hanja royal records (Hj$^{\text{R}}$), Hanja literary works (Hj$^{\text{L}}$), and Classical Chinese (Lzh), with error bars of 95\% confidence intervals for MT and standard deviations for NER and PR. Statistical significance is denoted as: *** ($p < 0.001$), ** ($p < 0.01$), * ($p < 0.05$), and n.s. (not significant).}
  \label{fig:mainrq}
\end{figure*}

%% file: contents/2_background.tex
Written languages in the Sinosphere initially adopted Classical Chinese syntax and vocabulary (Figure~\ref{fig:teaser_image}), but gradually diverged over time to meet local needs~\citep{handel-2019-sinography}.
This linguistic evolution has led to differences that potentially affect the efficacy of cross-lingual transfer in NLP tasks.
First, several characters became archaic, were transformed, and substituted by preferred heteromorphic synonyms, as Classical Chinese was disseminated into neighboring countries~\citep{kim-2012-hanja}.
Table~\ref{tab:varform} illustrates examples of regional variants between languages based on Classical Chinese.
Furthermore, Korea, Japan, and Vietnam developed variant forms and new characters to express local concepts~\citep{heo-2019-from}.
For instance, Koreans invented a new character 畓 (paddy field) in Hanja to reflect their agricultural lifestyle by combining two existing characters: 水 (water) and 田 (field).
Structural adaptations also occurred; while Classical Chinese typically follows a Subject-Verb-Object (SVO) structure, Kanbun adapted to a Subject-Object-Verb (SOV) structure, aligning more closely with Japanese grammar \citep{wang-etal-2023-kanbun}.

\input{tables/varform}

%% file: tables/varform.tex
\newcolumntype{C}{>{\centering\arraybackslash}X}
\begin{table}[t!]
  \centering
  \resizebox{\columnwidth}{!}{%
    \begin{tabularx}{\columnwidth}{@{}cCCCC@{}}
      \toprule
      \multicolumn{5}{c}{(a) Variant forms with same meaning}                                                                                                                                                                                                                                                                                         \\ \midrule
      \multirow{2}{*}{\textbf{Meaning}} & \multicolumn{4}{c}{\textbf{Preferred Form}}                                                                                                                                                                                                                                                                 \\ \cmidrule(l){2-5}
      & \textbf{CN}                                                                                                                                                                                                                                                       & \textbf{KR} & \textbf{JP} & \textbf{VN} \\ \midrule
      fight                             & \cjk{鬥}                                                                                                                                                                                                                                                           & 鬪           & \cjkj{闘}    & \cjk{鬥}     \\
      truly                             & \cjk{真}                                                                                                                                                                                                                                                           & 眞           & \cjk{真}     & \cjk{真}     \\
      leg                               & \cjk{腳}                                                                                                                                                                                                                                                           & 脚           & 脚           & \cjk{腳}     \\ \midrule
      \multicolumn{5}{c}{(b) Homographs with different meanings}                                                                                                                                                                                                                                                                                      \\ \midrule
      \multirow{2}{*}{\textbf{Char.}}   & \multicolumn{4}{c}{\textbf{Primary Meaning}}                                                                                                                                                                                                                                                                \\ \cmidrule(l){2-5}
      & \textbf{CN}                                                                                                                                                                                                                                                       & \textbf{KR} & \textbf{JP} & \textbf{VN} \\ \midrule
      空                                 & in vain                                                                                                                                                                                                                                                           & empty       & empty       & without     \\
      骨                                 & bone                                                                                                                                                                                                                                                              & bone        & cremains    & pillar      \\
      串                                 & skewer                                                                                                                                                                                                                                                            & cape        & skewer      & skewer      \\ \midrule
      \multicolumn{5}{c}{(c) Locally invented characters}                                                                                                                                                                                                                                                                                             \\ \midrule
      \textbf{Loc.}                     & \multicolumn{4}{c}{\textbf{Characters}}                                                                                                                                                                                                                                                                     \\ \midrule
      KR                                & \multicolumn{4}{l}{畓 (paddy field), 欌 (wardrobe)}                                                                                                                                                                                                                                                           \\
      JP                                & \multicolumn{4}{l}{\cjkj{榊} (sakaki tree), \cjkj{働} (work)}                                                                                                                                                                                                                                                 \\
      VN                                & \multicolumn{4}{l}{\raisebox{-0.2\height}{\includegraphics[height=1em]{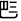}} (three), \raisebox{-0.2\height}{\includegraphics[height=1em]{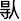}} (human), \raisebox{-0.2\height}{\includegraphics[height=1em]{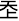}} (sky)}                                           \\
      \bottomrule
    \end{tabularx}%
  }
  \caption{Linguistic divergence patterns in the Sinosphere writing systems. It illustrates three types of character variations across China (CN), Korea (KR), Japan (JP), and Vietnam (VN): variant forms sharing meanings, homographs with distinct regional interpretations, and locally invented characters.}
  \label{tab:varform}
\end{table}

%% file: contents/3_experiments.tex
In this section, we detail the design, implementation, and results of our experiments investigating the impact of using Classical Chinese datasets to train language models for ancient Korean documents written in Hanja.

\subsection{Study Design}

\subsubsection{Documents}

\input{tables/data_statistics}

We construct our dataset by gathering publicly available resources and datasets written in languages within the Sinosphere.
To the best of our knowledge, resources for Kanbun and \chuhan{} are severely limited; small sizes of raw corpora exist for both, with some partial translations available for Kanbun.
Therefore, we focus on Hanja (Hj) and Classical Chinese (Lzh) for our experiments.
Hanja documents are further divided into two categories based on authorship: historical records written by government offices of the Joseon Dynasty (Hj$^{\text{R}}$) and literary work written by individual scholars (Hj$^{\text{L}}$).
Table~\ref{tab:data_stats} lists these corpora with their statistics.
See Appendix~\ref{sec:appx_replication_details} for more details, including data sources and preprocessing procedures.

\paragraph{Royal Documents in Hanja (Hj$^{\text{R}}$)} consists of government-compiled chronicles from the Joseon Dynasty period: \emph{the Annals of the Joseon Dynasty} (AJD), \emph{the Diaries of the Royal Secretariat} (DRS), and \emph{the Daily Records of the Royal Court and Important Officials} (DRRI).
These documents follow strict writing guidelines and exhibit a highly consistent style.

\paragraph{Literary Documents in Hanja (Hj$^{\text{L}}$)} refers to literary works written in Hanja authored by various Korean authors.
In this paper, we use \emph{the Korean Literary Collections} (KLC)\thinspace\footnote{also known as \emph{the Comprehensive Publication of Korean Literary Collections in Classical Chinese}} as the primary source.
Hanja literary works remain understudied in the NLP community, and the KLC corpus has not previously been explored in NLP research.
Detailed documentation of the KLC dataset is provided in Appendix~\ref{sec:appx_klc_dataset}.

\paragraph{Documents in Classical Chinese (Lzh)} comprises the WYWEB benchmark \citep{zhou-etal-2023-wyweb}, the NiuTrans Classical Chinese to Modern Chinese dataset\thinspace\footnote{\ztexttt{\url{https://github.com/NiuTrans/Classical-Modern}}}, the C2MChn dataset \citep{jiang2023c2mchn}, Daizhige\thinspace\footnote{\ztexttt{\url{https://github.com/garychowcmu/daizhigev20}}}, and the Oriental Classics Database (OCDB)\thinspace\footnote{\ztexttt{\url{http://db.cyberseodang.or.kr}}}.
WYWEB consists of nine NLP tasks for Classical Chinese, including GLNER---a named entity recognition task initially developed by \citet{gulian-2020}---and WYWMT---a machine translation task that translates Classical Chinese into Modern Chinese.
Daizhige, the largest classical Chinese corpus, contains about 2.4 billion tokens of classical literature.
The OCDB provides original Chinese texts and Korean translations of authoritative books.

\paragraph{Other Documents in the Sinosphere.} We collect historical documents from Japan and Vietnam and analyze them in the discussion section.
For Kanbun, we use the \emph{Rikkokushi}, Japan's Six National Histories.
For Chữ Hán, we include four major Vietnamese historical chronicles: the \emph{Đại Việt sử ký toàn thư} (ĐVSKTT) and \emph{Đại Nam thực lục} (ĐNTL), which served as official dynastic records, along with the \emph{An Nam chí lược} (ANCL) and \emph{Đại Việt sử lược} (ĐVSL).

\paragraph{Data Augmentation.} We create a synthetic dataset that translates Classical Chinese into Korean by applying machine translation to Modern Chinese sentences from the NiuTrans dataset.
Translation efforts for Classical Chinese predominantly focus on Modern Chinese, making it challenging to explore cross-lingual transferability.
We employ GPT-4\thinspace\footnote{The experiments were conducted on April 6, 2024 -- April 12, 2024 with \ztexttt{gpt-4-0125-preview} model under Azure OpenAI Service with the OpenAI API as a fallback when content filtering prevented response generation.} to generate a total of 972,467 synthetic sentence pairs from Classical Chinese to Korean, adapting the approach proposed by \citet{nehrdich-etal-2023-mitra}.
Detailed inference settings are provided in Appendix~\ref{sec:appx_data_aug}.

\subsubsection{Tasks}

The experiments focus on three tasks: machine translation (MT), named entity recognition (NER), and punctuation restoration (PR).
These tasks represent real-world challenges for human experts analyzing and understanding ancient languages.

\paragraph{Machine Translation (MT)} of ancient Korean documents into modern languages is crucial, as most contemporary Koreans, including scholars, cannot comprehend Hanja texts without translation.
We measure the BLEU score~\citep{papineni-etal-2002-bleu} using SacreBLEU~\citep{post-2018-call}.

\paragraph{Named Entity Recognition (NER)} is a sequence labeling task that identifies and classifies proper names, such as persons and locations, in text.
Combined with entity linking, it is crucial for indexing and searching large historical records.
We report the F1-score after normalizing all predicted and ground-truth labels to `NE', akin to the binary setting in NLTK, to ensure a fair comparison across different models and datasets.
For readability, F1-scores are presented as percentages (0-100) in tables and figures, while being expressed in the standard 0-1 scale in the text (\eg 87.5 = 0.875).

\paragraph{Punctuation Restoration (PR)} is an essential pre-translation step that involves inserting modern punctuation marks into original Hanja texts, as punctuation greatly impacts the meaning of these texts.
We adopt the comprehensive punctuation restoration approach proposed by \citet{pogoda-walkowiak-2021-comprehensive-punctuation} for training.
For evaluation, we use the weighted average F1-score after simplifying each punctuation combination to the conventionally defined 4-class task (comma, period, question mark, and other).
Reduction rules are presented in Appendix~\ref{sec:appx_inference_evaluation}.

\subsubsection{Model Training}

We fine-tune Qwen2-7B~\citep{yang-2024-qwen2} for MT and SikuRoBERTa~\citep{sikubert-2021} for NER and PR, respectively.
Table~\ref{tab:train_data_composition} in Appendix~\ref{sec:appx_exp_setup} presents the composition of training data for each task.
For documents without predefined splits, we allocate 80\% for training, 10\% for validation, and 10\% for testing.
The KLC data is bifurcated at the book level for training/validation and testing.

\paragraph{Qwen2} is a series of foundation models pre-trained on multilingual corpus and proficient in over 30 languages, including Chinese, Korean, and English~\citep{yang-2024-qwen2}.
We fine-tune the Qwen2-7B using QLoRA~\citep{dettmers-2023-qlora} for machine translation of three language pairs: Hj-Ko, Hj-En, and Lzh-Ko, using the prompt in Appendix~\ref{sec:appx_training_hparams}.

\paragraph{SikuRoBERTa} is a RoBERTa-based model pretrained on the \emph{Siku Quanshu}, a vast collection of Classical Chinese literature \citep{sikubert-2021}.\footnote{Encoder-based models pretrained on Classical Chinese corpora have been employed by multiple Hanja-related studies \citep{yoo-etal-2022-hue, moon-2024-exploiting}.}

\input{tables/main_result}

\subsection{Experimental Results}
\label{sec:exp_results}

We evaluate models trained across various dataset combinations and tasks, with results shown in Table~\ref{tab:main_result}.
Incorporating Classical Chinese resources yields minimal or non-significant improvements for Hanja documents across all tasks.
For machine translation, significance testing via paired bootstrap resampling \citep{koehn-2004-statistical} reveals that only 2 of 9 test conditions show improvements.
The largest gain (+1.01 BLEU for Hj$^{\text{L}}$-Ko) achieves only 60-65\% agreement with human judgments~\citep{kocmi-etal-2024-navigating}, while most conditions show decreases or stagnation (-3.14 to +0.84 BLEU).
For sequence labeling tasks (\ie NER and PR), 5-fold cross-validation with Mann-Whitney $U$ tests \citep{mann-1947-test} shows no significant changes ($p<0.05$) when adding Classical Chinese data, with F1-score differences ranging from -0.0215 to +0.0067.
In contrast, Classical Chinese documents show significant performance improvements when trained with Classical Chinese resources, indicating successful baseline training.
A qualitative error analysis of these results is available in Appendix~\ref{sec:appx_error_analysis}.

Notably, models trained exclusively on Classical Chinese perform well on sequence labeling tasks for Hanja documents, with the Classical Chinese NER model outperforming Hj$^{\text{R}}$-trained model on Hj$^{\text{L}}$ data (0.7261 vs 0.7082 F1).
While machine translation requires comprehensive language understanding and generation capabilities, NER and PR primarily capture character and word-level patterns.
The smaller performance variations in PR compared to MT and NER suggest that punctuation patterns exhibit a degree of consistency across the Sinosphere writing systems.

Our results reveal a clear division between royal and literary Hanja texts.
Models trained on Hj$^{\text{R}}$ perform poorly on Hj$^{\text{L}}$ (BLEU scores below 11.82), with similar patterns in NER.
This aligns with known linguistic differences between government chronicles, which follow strict guidelines, and diverse literary works by individual authors \citep{moon-2024-exploiting}.

For Classical Chinese language modeling, incorporating Hanja data shows minimal impact.
Adding Hj$^{\text{L}}$ produces no significant changes across tasks, while Hj$^{\text{R}}$ data yields modest differences (+0.50 BLEU, +0.0137 F1, and -0.0058 F1 for MT, NER, and PR, respectively).

%% file: tables/data_statistics.tex
\begin{table*}[t!]
  \centering
  \resizebox{\textwidth}{!}{%
    \begin{tabular}{@{}cllccccrrrr@{}}
      \toprule
      \multirow{2}{*}{\textbf{Language}}                      &
      \multirow{2}{*}{\textbf{Type}}                          &
      \multirow{2}{*}{\textbf{Document}}                      &
      \multirow{2}{*}{\textbf{Time Period}}                   &
      \multicolumn{3}{c}{\textbf{Tasks}}                      &
      \multirow{2}{*}{\textbf{\# of Samples}}                 &
      \multirow{2}{*}{\textbf{
          \begin{tabular}[c]{@{}r@{}}Avg. \# of \\ Characters
      \end{tabular}}}     &
      \multirow{2}{*}{\textbf{
          \begin{tabular}[c]{@{}r@{}}\# of Tokens \\ (GPT-4)
      \end{tabular}}}      &
      \multirow{2}{*}{\textbf{
          \begin{tabular}[c]{@{}r@{}}Trans. \\ (\%)
      \end{tabular}}}                  \\
      &
      &
      &
      &
      \textbf{MT}                                             &
      \textbf{NER}                                            &
      \textbf{PR}                                             &
      &
      &
      &
      \\ \midrule
      \multirow{4}{*}{
        \begin{tabular}[c]{@{}c@{}}Hanja \\ (Hj)
      \end{tabular}}                &
      \multirow{3}{*}{Royal}                                  &
      AJD                                                     &
      1392-1928                                               &
      \ding{52}                                               &
      \ding{52}                                               &
      \ding{52}                                               &
      413,323                                                 &
      173.9                                                   &
      103,013,789                                             &
      100.0                                                     \\
      &
      &
      DRS                                                     &
      1623-1910                                               &
      \ding{52}                                               &
      -                                                       &
      -                                                       &
      1,787,007                                               &
      165.2                                                   &
      433,873,833                                             &
      30.9                                                      \\
      &
      &
      DRRI                                                    &
      1760-1910                                               &
      \ding{52}                                               &
      -                                                       &
      -                                                       &
      616,910                                                 &
      81.1                                                    &
      84,141,022                                              &
      32.6                                                      \\
      &
      Literary                                                &
      KLC                                                     &
      886-1933                                                &
      \ding{52}                                               &
      \ding{52}                                               &
      \ding{52}                                               &
      653,386                                                 &
      336.7                                                   &
      340,113,975                                             &
      29.8                                                      \\ \midrule
      \multirow{7}{*}{
        \begin{tabular}[c]{@{}c@{}}Classical \\ Chinese\\ (Lzh)
      \end{tabular}} &
      \multirow{7}{*}{Mixed}                                  &
      Daizhige\textsuperscript{$\dagger$}                     &
      -                                                       &
      -                                                       &
      -                                                       &
      -                                                       &
      15,694                                                  &
      107,636.9                                               &
      2,449,254,631                                           &
      -                                                         \\
      &
      &
      NiuTrans                                                &
      -                                                       &
      \ding{52}                                               &
      -                                                       &
      -                                                       &
      972,467                                                 &
      22.4                                                    &
      31,312,241                                              &
      100.0                                                     \\
      &
      &
      C2MChn\textsuperscript{$\dagger$}                       &
      -                                                       &
      \ding{52}                                               &
      -                                                       &
      -                                                       &
      614,723                                                 &
      18.9                                                    &
      17,845,525                                              &
      100.0                                                     \\
      &
      &
      OCDB                                                    &
      6 c. BC-16 c.                                           &
      \ding{52}                                               &
      -                                                       &
      -                                                       &
      23,795                                                  &
      230.9                                                   &
      8,018,473                                               &
      100.0                                                     \\
      &
      &
      WYWMT                                                   &
      -                                                       &
      \ding{52}                                               &
      -                                                       &
      -                                                       &
      266,514                                                 &
      21.9                                                    &
      8,293,026                                               &
      100.0                                                     \\
      &
      &
      GLNER                                                   &
      -                                                       &
      -                                                       &
      \ding{52}                                               &
      -                                                       &
      18,762                                                  &
      209.7                                                   &
      5,416,667                                               &
      -                                                         \\
      &
      &
      WYWEB                                                   &
      1046 BC-1927                                            &
      -                                                       &
      -                                                       &
      \ding{52}                                               &
      135,134                                                 &
      117.5                                                   &
      22,753,344                                              &
      -                                                         \\ \midrule
      Kanbun (Kb)                                             &
      Royal                                                   &
      Rikkokushi\textsuperscript{$\dagger$}                   &
      697-887                                                 &
      \ding{52}                                               &
      -                                                       &
      -                                                       &
      17,306                                                  &
      83.5                                                    &
      2,291,164                                               &
      9.1                                                       \\ \midrule
      \multirow{4}{*}{\chuhan{}}                              &
      \multirow{4}{*}{Royal}                                  &
      ĐVSKTT\textsuperscript{$\dagger$}                       &
      2 c. BC-1675                                            &
      -                                                       &
      -                                                       &
      -                                                       &
      8,484                                                   &
      52.4                                                    &
      872,620                                                 &
      -                                                         \\
      &
      &
      ĐNTL\textsuperscript{$\dagger$}                         &
      1545-1909                                               &
      -                                                       &
      -                                                       &
      -                                                       &
      5,608                                                   &
      58.8                                                    &
      475,523                                                 &
      -                                                         \\
      &
      &
      ANCL\textsuperscript{$\dagger$}                         &
      1285-1339                                               &
      -                                                       &
      -                                                       &
      -                                                       &
      1,288                                                   &
      65.3                                                    &
      135,159                                                 &
      -                                                         \\
      &
      &
      ĐVSL\textsuperscript{$\dagger$}                         &
      2 c. BC-1225                                            &
      -                                                       &
      -                                                       &
      -                                                       &
      1,164                                                   &
      66.3                                                    &
      63,677                                                  &
      -                                                         \\ \bottomrule
    \end{tabular}%
  }
  \caption{Statistics of historical documents from the Sinosphere. Documents marked with $\dagger$ are supplementary materials analyzed in discussions and not used in the main experimental evaluations. Trans. (\%) indicates the ratio of documents with publicly available translations, and tokens are counted using tiktoken's \ztexttt{cl100k\_base} encoding.}
  \label{tab:data_stats}
\end{table*}

%% file: tables/main_result.tex
\begin{table}[t!]
  \centering
  \resizebox{\linewidth}{!}{%
    \begin{tabular}{@{}ccccccc@{}}
      \toprule
      \multicolumn{7}{c}{(a) Machine Translation (MT)}                                                                                                                                                                                               \\ \midrule
      \multicolumn{3}{c}{\textbf{Train Data}} & \multicolumn{4}{c}{\textbf{Test Data (BLEU)}}                                                                                                                                                        \\
      \textbf{Hj$^{\text{R}}$}                & \textbf{Hj$^{\text{L}}$}                          & \textbf{Lzh} & \textbf{Hj$^{\text{R}}$-En}                  & \textbf{Hj$^{\text{R}}$-Ko} & \textbf{Hj$^{\text{L}}$-Ko} & \textbf{Lzh-Ko}        \\ \midrule
      &                                                   & \ding{52}    & 0.02                                         & 9.79                        & 4.85                        & 18.13                  \\ \hdashline
      \ding{52}                               &                                                   &              & \textbf{33.16}                               & {\ul 47.93}                 & 10.81                       & 11.64                  \\
      \ding{52}                               &                                                   & \ding{52}    & 31.34                                        & 47.17                       & 11.82                       & {\ul 18.63}            \\
      &                                                   &              & \cb{ffe8bc}{($-1.82$)}                       & \cb{fff5e3}{($-0.76$)}      & \cb{daf2ff}{($+1.01$)}      & \cb{00a9ff}{($+6.99$)} \\ \hdashline
      & \ding{52}                                         &              & 0.13                                         & 34.16                       & {\ul 33.57}                 & 11.91                  \\
      & \ding{52}                                         & \ding{52}    & 0.06                                         & 31.02                       & 32.19                       & 18.06                  \\
      &                                                   &              & \notsig ($-0.07$)                            & \cb{ffd88c}{($-3.14$)}      & \cb{ffeecc}{($-1.38$)}      & \cb{1eb4ff}{($+6.15$)} \\ \hdashline
      \ding{52}                               & \ding{52}                                         &              & {\ul 33.15}                                  & \textbf{48.97}              & 33.07                       & 12.32                  \\
      \ding{52}                               & \ding{52}                                         & \ding{52}    & 31.52                                        & 47.49                       & \textbf{33.91}              & \textbf{18.78}         \\
      &                                                   &              & \cb{ffebc3}{($-1.63$)}                       & \cb{ffedc9}{($-1.48$)}      & \cb{e0f4ff}{($+0.84$)}      & \cb{13b0ff}{($+6.46$)} \\ \midrule
      \multicolumn{7}{c}{(b) Named Entity Recognition (NER)}                                                                                                                                                                                         \\ \midrule
      \multicolumn{3}{c}{\textbf{Train Data}} & \multicolumn{4}{c}{\textbf{Test Data (F1-score)}}                                                                                                                                                    \\
      \textbf{Hj$^{\text{R}}$}                & \textbf{Hj$^{\text{L}}$}                          & \textbf{Lzh} & \multicolumn{2}{c}{\textbf{Hj$^{\text{R}}$}} & \textbf{Hj$^{\text{L}}$}    & \textbf{Lzh}                                         \\ \midrule
      &                                                   & \ding{52}    & \multicolumn{2}{c}{81.32}                    & 72.61                       & 86.48                                                \\ \hdashline
      \ding{52}                               &                                                   &              & \multicolumn{2}{c}{{\ul 97.51}}              & 70.82                       & 65.15                                                \\
      \ding{52}                               &                                                   & \ding{52}    & \multicolumn{2}{c}{97.47}                    & 70.01                       & \textbf{87.85}                                       \\
      &                                                   &              & \multicolumn{2}{c}{\cb{e0e0e0}{($-0.04$)}}   & \notsig ($-0.81$)           & \cb{00a9ff}{($+22.70$)}                              \\ \hdashline
      & \ding{52}                                         &              & \multicolumn{2}{c}{88.99}                    & {\ul 83.63}                 & 66.31                                                \\
      & \ding{52}                                         & \ding{52}    & \multicolumn{2}{c}{86.84}                    & 83.13                       & 87.05                                                \\
      &                                                   &              & \multicolumn{2}{c}{\cb{fff6e6}{($-2.15$)}}   & \notsig ($-0.50$)           & \cb{16b1ff}{($+20.74$)}                              \\ \hdashline
      \ding{52}                               & \ding{52}                                         &              & \multicolumn{2}{c}{\textbf{97.53}}           & 83.55                       & 66.15                                                \\
      \ding{52}                               & \ding{52}                                         & \ding{52}    & \multicolumn{2}{c}{97.45}                    & \textbf{84.22}              & {\ul 87.68}                                          \\
      &                                                   &              & \multicolumn{2}{c}{\cb{fffefe}{($-0.08$)}}   & \notsig ($+0.67$)           & \cb{0daeff}{($+21.53$)}                              \\ \midrule
      \multicolumn{7}{c}{(c) Punctuation Restoration (PR)}                                                                                                                                                                                           \\ \midrule
      \multicolumn{3}{c}{\textbf{Train Data}} & \multicolumn{4}{c}{\textbf{Test Data (F1-score)}}                                                                                                                                                    \\
      \textbf{Hj$^{\text{R}}$}                & \textbf{Hj$^{\text{L}}$}                          & \textbf{Lzh} & \multicolumn{2}{c}{\textbf{Hj$^{\text{R}}$}} & \textbf{Hj$^{\text{L}}$}    & \textbf{Lzh}                                         \\ \midrule
      &                                                   & \ding{52}    & \multicolumn{2}{c}{78.36}                    & 80.66                       & {\ul 85.83}                                          \\ \hdashline
      \ding{52}                               &                                                   &              & \multicolumn{2}{c}{88.58}                    & 84.77                       & 77.25                                                \\
      \ding{52}                               &                                                   & \ding{52}    & \multicolumn{2}{c}{{\ul 88.60}}              & 84.61                       & 85.25                                                \\
      &                                                   &              & \multicolumn{2}{c}{\cb{e0e0e0}{($+0.02$)}}   & \notsig ($-0.16$)           & \cb{00a9ff}{($+8.00$)}                               \\ \hdashline
      & \ding{52}                                         &              & \multicolumn{2}{c}{80.49}                    & 87.05                       & 79.45                                                \\
      & \ding{52}                                         & \ding{52}    & \multicolumn{2}{c}{80.66}                    & 87.27                       & \textbf{85.95}                                       \\
      &                                                   &              & \multicolumn{2}{c}{\cb{e0e0e0}{($+0.17$)}}   & \notsig ($+0.22$)           & \cb{2fb9ff}{($+6.50$)}                               \\ \hdashline
      \ding{52}                               & \ding{52}                                         &              & \multicolumn{2}{c}{\textbf{88.61}}           & {\ul 87.76}                 & 78.02                                                \\
      \ding{52}                               & \ding{52}                                         & \ding{52}    & \multicolumn{2}{c}{88.57}                    & \textbf{87.91}              & 85.28                                                \\
      &                                                   &              & \multicolumn{2}{c}{\cb{e0e0e0}{($-0.04$)}}   & \notsig ($+0.15$)           & \cb{17b1ff}{($+7.26$)}                               \\
      \bottomrule
    \end{tabular}
  }
  \caption{Performance comparisons for MT, NER, and PR tasks across all combinations of document types used in training.
    The values in parentheses denote the score differences between the models trained with and without Classical Chinese data (Lzh).
    \colorbox{lightgray!100}{Gray} indicates no significant differences.
    \cb{ffaa00}{Orange} and \cb{00aaff}{blue} indicate significant decreases and increases, respectively, with saturation reflecting the magnitude of differences by each task.
    \textbf{Bold} and \underline{underlined} numbers denote the highest and the second-highest scores for each task and test dataset, respectively.
  }
  \label{tab:main_result}
\end{table}

%% file: contents/4_discussion.tex
In this section, we explore potential reasons why Classical Chinese exhibits limited impact on the language models for Asian historical documents and support them with empirical analyses.

\subsection{Model Scaling and Architecture Variations}
\label{sec:model-scaling}

\input{tables/model_scaling}
\input{tables/model_arch}

We extend our observations to smaller model scales (Table~\ref{tab:model_scaling}) and various foundation models (Table~\ref{tab:model_arch}) by fine-tuning MT models with and without Classical Chinese data.
We outline that incorporating Classical Chinese corpora significantly impairs Hanja language modeling across both smaller scales of Qwen2 and different foundation models (\ie Llama-3.1-8B-Instruct and Gemma-2-9B).
Specifically, BLEU scores for Hanja-to-English and Hanja-to-Korean on royal documents decrease by 5.08 and 5.94, respectively, when fine-tuning Qwen2-1.5B.

\subsection{Threshold for Diminishing Benefits of Classical Chinese Data}
\label{sec:threshold-diminishing}

We hypothesize that sufficient Hanja data exists to train effective language models without relying on Classical Chinese resources, given the substantial volume of annotated Hanja documents preserved through national research initiatives.
When measured by token count, available training data for Hanja exceeds Classical Chinese by factors of 4.4, 18.6, and 6.8 for MT, NER, and PR, respectively.

\input{figures/low-resource}

To identify the threshold where Classical Chinese data ceases to provide meaningful benefits, we conduct an ablation study by systematically varying the ratio of Hanja to Classical Chinese training data.
Figure~\ref{fig:low-resource} shows performance differences between models trained with and without Classical Chinese data across different Hanja data proportions.
While Classical Chinese resources significantly boost performance in extremely low-resource scenarios, particularly for literary documents, these benefits diminish rapidly as Hanja data increases.
The performance improvements become relatively small (below 5.5\%) across all tasks once Hanja data exceeds one-eighth the volume of Classical Chinese data.
Detailed results are in Table~\ref{tab:low-resource_all}.
These findings suggest that while Classical Chinese resources can be valuable in low-resource settings, their utility diminishes quickly with increasing Hanja data availability, challenging the assumption that incorporating additional auxiliary data consistently improves performance.

\subsection{Domain-Specific Transfer Learning}
\label{sec:domain-specific}

We further investigate whether targeting specific domains of Classical Chinese data can improve cross-lingual transfer effectiveness for Hanja.
Using the C2MChn dataset~\citep{jiang2023c2mchn}, we categorize Classical Chinese texts into three domains aligned with Hanja genres: History, Religion (Buddhism, Confucianism, Taoism), and Miscellaneous (Agronomy, Short, Others), and conduct fine-tuning experiments with Qwen2-7B using various domain combinations.

\input{tables/domain_specific}

Table~\ref{tab:domain-specific} shows that incorporating Classical Chinese data from any domain combination reduces MT model performance for Hanja royal documents compared to using Hanja data alone.
While the Miscellaneous domain occasionally produces minor improvements for literary documents (maximum +1.41 BLEU), the overall effects remain mixed or negligible.
We hypothesize that short-form poetry within the Miscellaneous domain may assist with similarly styled Hanja literary works, but using untargeted data across domains diminishes this benefit.
These results underscore that domain-specific Classical Chinese data requires careful empirical validation for effective use.

\subsection{Expandability to Sinosphere}
\label{sec:expandability-sinosphere}

\subsubsection{Machine Translation for Kanbun}
\label{sec:mt-kanbun}

\input{tables/kanbun-mt}

To explore the generalizability of our findings to other languages in the Sinosphere, we conduct experiments on Kanbun using 1,371 paragraph-level samples from Korean-related records\thinspace\footnote{\ztexttt{\url{https://db.history.go.kr/id/jm}}} in the Six National Histories of Japan.
As shown in Table~\ref{tab:kanbun-mt}, both Hanja and Classical Chinese resources improve Kanbun translation performance (BLEU scores increase by 19.17 and 11.14, respectively), demonstrating that cross-lingual transfer can be effective in low-resource settings.
However, careful empirical validation is needed when selecting source languages rather than simply combining all available resources.

Here, the varying degrees of improvement likely stem from different levels of linguistic and topical similarity.
We validate this empirically using 5-gram language models trained on Korean translations, where perplexity on Kanbun documents is lower with a model trained on Hanja (181) versus Classical Chinese (264).
This pattern reflects our test set composition: Korea-related Kanbun texts translated by a Korean institution.

\subsubsection{Vocabulary Divergence}
\label{sec:vocab-divergence}

We computationally identify the linguistic distance between Classical Chinese and other writing systems in the Sinosphere through character-level analysis.
Analysis of unique characters across writing systems (Figure~\ref{fig:vocab}) reveals Hanja having the largest vocabulary (23,186 characters), followed by Classical Chinese, \chuhan{}, and Kanbun.
While 32.2\% of Hanja characters do not appear in our Classical Chinese corpus, these Hanja-exclusive characters occur infrequently, comprising less than 1.9\% of character usage at the 99\% frequency threshold (Figure~\ref{fig:vocab-heatmap}).
Further inspection reveals that most Hanja-exclusive characters are documented variant forms of Classical Chinese characters in the \emph{Kangxi Dictionary}, rather than Korean-invented characters.
For instance, the character {\includegraphics[height=0.9em]{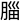}} in the Annals of the Joseon Dynasty is a known variant of \cjk{腦} (brain) but absent from our Classical Chinese corpora.
While variant character normalization techniques \citep{kessler-2024-towards} might mitigate these surface-level differences, our findings suggest that the challenges in cross-lingual transfer stem from factors beyond vocabulary divergence.

\input{figures/vocab}

\input{figures/vocab-heatmap}

%% file: tables/model_scaling.tex
\begin{table}[t]
  \centering
  \resizebox{\columnwidth}{!}{%
    \begin{tabular}{@{}ccccccc@{}}
      \toprule
      \multirow{2}{*}{\textbf{
          \begin{tabular}[c]{@{}c@{}}Model \\ Size
      \end{tabular}}}     &
      \multicolumn{2}{c}{\textbf{Train}}           &
      \multirow{2}{*}{\textbf{Hj$^{\text{R}}$-En}} &
      \multirow{2}{*}{\textbf{Hj$^{\text{R}}$-Ko}} &
      \multirow{2}{*}{\textbf{Hj$^{\text{L}}$-Ko}} &
      \multirow{2}{*}{\textbf{Lzh-Ko}}                                                                                                                                              \\
      & \textbf{Hj} & \textbf{Lzh} &                        &                        &                        &                        \\ \midrule
      \multirow{3}{*}{7B}                          & \ding{52}   &              & \textbf{33.15}         & \textbf{48.97}         & {\ul 33.07}            & 12.32                  \\
      & \ding{52}   & \ding{52}    & {\ul 31.52}            & {\ul 47.49}            & \textbf{33.91}         & \textbf{18.78}         \\
      &             &              & \cb{ffeac1}{($-1.63$)} & \cb{ffecc6}{($-1.48$)} & \cb{dff4ff}{($+0.84$)} & \cb{09adff}{($+6.46$)} \\ \hdashline
      \multirow{3}{*}{1.5B}                        & \ding{52}   &              & 28.74                  & 43.58                  & 29.32                  & 8.92                   \\
      & \ding{52}   & \ding{52}    & 23.66                  & 37.64                  & 26.66                  & {\ul 15.61}            \\
      &             &              & \cb{ffbe3e}{($-5.08$)} & \cb{ffb31d}{($-5.94$)} & \cb{ffdd9a}{($-2.66$)} & \cb{01aaff}{($+6.69$)} \\ \hdashline
      \multirow{3}{*}{0.5B}                        & \ding{52}   &              & 17.34                  & 34.14                  & 21.30                  & 3.45                   \\
      & \ding{52}   & \ding{52}    & 14.38                  & 33.01                  & 16.77                  & 10.17                  \\
      &             &              & \cb{ffd98e}{($-2.96$)} & \cb{fff0d4}{($-1.13$)} & \cb{ffc553}{($-4.53$)} & \cb{00a9ff}{($+6.72$)} \\ \bottomrule
    \end{tabular}%
  }
  \caption{BLEU scores of machine translation models at varying parameter scales trained with/without Classical Chinese (Lzh) data.
  }
  \label{tab:model_scaling}
\end{table}

%% file: tables/model_arch.tex
\begin{table}[t]
  \centering
  \resizebox{\columnwidth}{!}{%
    \begin{tabular}{@{}ccccccc@{}}
      \toprule
      \multirow{2}{*}{\textbf{Model}}              &
      \multicolumn{2}{c}{\textbf{Train}}           &
      \multirow{2}{*}{\textbf{Hj$^{\text{R}}$-En}} &
      \multirow{2}{*}{\textbf{Hj$^{\text{R}}$-Ko}} &
      \multirow{2}{*}{\textbf{Hj$^{\text{L}}$-Ko}} &
      \multirow{2}{*}{\textbf{Lzh-Ko}}                                                                                                                                              \\
      & \textbf{Hj} & \textbf{Lzh} &                        &                        &                        &                        \\ \midrule
      \multirow{3}{*}{Qwen2}                       & \ding{52}   &              & 33.15                  & 48.97                  & 33.07                  & 12.32                  \\
      & \ding{52}   & \ding{52}    & 31.52                  & 47.49                  & 33.91                  & {\ul 18.78}            \\
      &             &              & \cb{ffe9be}{($-1.63$)} & \cb{ffebc4}{($-1.48$)} & \cb{ddf3ff}{($+0.84$)} & \cb{00a9ff}{($+6.46$)} \\ \hdashline
      \multirow{3}{*}{Llama-3.1}                   & \ding{52}   &              & {\ul 33.96}            & 49.03                  & 34.56                  & 13.13                  \\
      & \ding{52}   & \ding{52}    & 32.25                  & 47.53                  & 33.50                  & 18.76                  \\
      &             &              & \cb{ffe8bb}{($-1.71$)} & \cb{ffebc3}{($-1.50$)} & \cb{fff1d5}{($-1.06$)} & \cb{20b4ff}{($+5.63$)} \\ \hdashline
      \multirow{3}{*}{Gemma-2}                     & \ding{52}   &              & \textbf{35.39}         & \textbf{51.86}         & \textbf{36.69}         & 13.20                  \\
      & \ding{52}   & \ding{52}    & 33.56                  & {\ul 49.66}            & {\ul 35.09}            & \textbf{19.61}         \\
      &             &              & \cb{ffe6b6}{($-1.83$)} & \cb{ffe2a8}{($-2.20$)} & \cb{ffe9bf}{($-1.60$)} & \cb{01aaff}{($+6.41$)} \\ \bottomrule
    \end{tabular}%
  }
  \caption{BLEU scores of machine translation models across different architectures with/without Classical Chinese (Lzh) training data.
  }
  \label{tab:model_arch}
\end{table}

%% file: figures/low-resource.tex
\begin{figure*}[t!]
  \centering
  \subfloat[\centering Machine Translation]{\includegraphics[width=.32\linewidth]{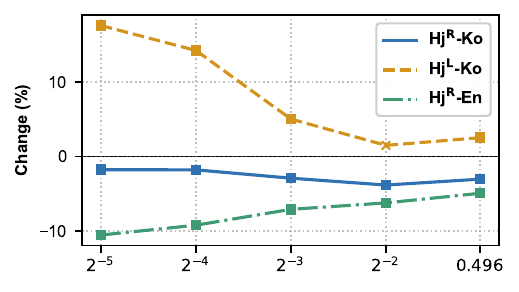}}\label{fig:low-resource_mt}
  \subfloat[\centering Named Entity Recognition]{\includegraphics[width=.32\linewidth]{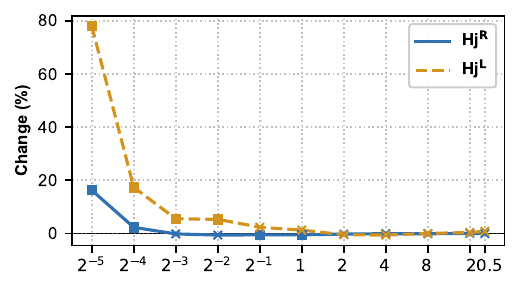}}\label{fig:low-resource_ner}
  \subfloat[\centering Punctuation Restoration]{\includegraphics[width=.32\linewidth]{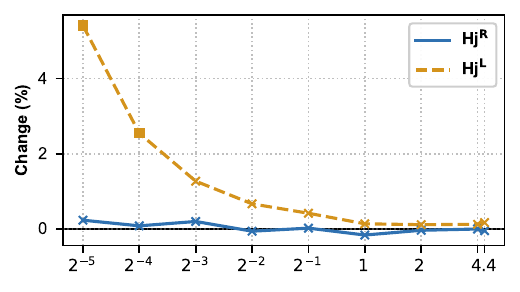}}\label{fig:low-resource_pr}
  \caption{Performance impact of Classical Chinese training data across varying Hanja data ratios. The $x$-axis shows the ratio $r$, where Hj:Lzh = $r$:1 denotes the proportion of Hanja data against Classical Chinese data, while the $y$-axis shows the relative performance differences in percentage (\%) between models trained with/without Classical Chinese data. Square and x markers indicate statistically significant differences ($p < 0.05$) and non-significant differences, respectively.}
  \label{fig:low-resource}
\end{figure*}

%% file: tables/domain_specific.tex
\begin{table}[t]
  \centering
  \resizebox{\columnwidth}{!}{%
    \begin{tabular}{@{}ccccccc@{}}
      \toprule
      \multicolumn{3}{c}{\textbf{Domain}}          &
      \multirow{2}{*}{\textbf{Hj$^{\text{R}}$-En}} &
      \multirow{2}{*}{\textbf{Hj$^{\text{R}}$-Ko}} &
      \multirow{2}{*}{\textbf{Hj$^{\text{L}}$-Ko}} &
      \multirow{2}{*}{\textbf{Lzh-Ko}}                                                                                                                                                   \\
      \textbf{His}                                 & \textbf{Rel}   & \textbf{Mis}   &                        &                        &                        &                        \\ \midrule
      \multicolumn{3}{c}{\textit{None (baseline)}} & \textbf{33.15} & \textbf{48.97} & 33.07                  & 12.32                                                                    \\ \midrule
      \ding{52}                                    &                &                & 32.26                  & 47.80                  & 33.60                  & 16.88                  \\
      &                &                & \cb{fff0d3}{($-0.89$)} & \cb{ffebc5}{($-1.17$)} & \cb{e0e0e0}{($+0.53$)} & \cb{1fb4ff}{($+4.56$)} \\ \hdashline
      & \ding{52}      &                & 32.23                  & 47.82                  & 33.68                  & 16.90                  \\
      &                &                & \cb{ffefd1}{($-0.92$)} & \cb{ffecc6}{($-1.15$)} & \cb{e0e0e0}{($+0.61$)} & \cb{1eb4ff}{($+4.58$)} \\ \hdashline
      &                & \ding{52}      & {\ul 32.71}            & {\ul 48.55}            & \textbf{34.48}         & 16.78                  \\
      &                &                & \cb{fff7e9}{($-0.44$)} & \cb{fff8ea}{($-0.42$)} & \cb{b9e7ff}{($+1.41$)} & \cb{24b6ff}{($+4.46$)} \\ \hdashline
      \ding{52}                                    & \ding{52}      &                & 31.98                  & 47.97                  & 32.27                  & \textbf{17.52}         \\
      &                &                & \cb{ffebc5}{($-1.17$)} & \cb{ffeecd}{($-1.00$)} & \cb{e0e0e0}{($-0.80$)} & \cb{00a9ff}{($+5.20$)} \\ \hdashline
      \ding{52}                                    &                & \ding{52}      & 31.89                  & 47.45                  & 34.03                  & 16.83                  \\
      &                &                & \cb{ffeac1}{($-1.26$)} & \cb{ffe6b4}{($-1.52$)} & \cb{cfefff}{($+0.96$)} & \cb{21b5ff}{($+4.51$)} \\ \hdashline
      & \ding{52}      & \ding{52}      & 31.80                  & 48.11                  & {\ul 34.06}            & 16.96                  \\
      &                &                & \cb{ffe8bc}{($-1.35$)} & \cb{fff0d4}{($-0.86$)} & \cb{ceeeff}{($+0.99$)} & \cb{1bb3ff}{($+4.64$)} \\ \hdashline
      \ding{52}                                    & \ding{52}      & \ding{52}      & 31.77                  & 47.37                  & 33.66                  & {\ul 17.47}            \\
      &                &                & \cb{ffe8bb}{($-1.38$)} & \cb{ffe4b0}{($-1.60$)} & \cb{e0e0e0}{($+0.59$)} & \cb{02aaff}{($+5.15$)} \\ \bottomrule
    \end{tabular}
  }
  \caption{Performance comparison of domain-specific transfer learning for machine translation. Models are trained on Hanja data (351.1M tokens) combined with different domains of Classical Chinese: History (23.6M tokens), Religion (21.6M tokens), and Miscellaneous (3.7M tokens).}
  \label{tab:domain-specific}
\end{table}

%% file: tables/kanbun-mt.tex
\begin{table}[t!]
  \centering
  \resizebox{\columnwidth}{!}{%
    \begin{tabular}{@{}ccccccc@{}}
      \toprule
      \multicolumn{3}{c}{\textbf{Train Data}}      &
      \multirow{2}{*}{\textbf{Kb-Ko}}              &
      \multirow{2}{*}{\textbf{Hj$^{\text{R}}$-Ko}} &
      \multirow{2}{*}{\textbf{Hj$^{\text{L}}$-Ko}} &
      \multirow{2}{*}{\textbf{Lzh-Ko}}                                                                                                              \\
      \textbf{Kb}                                  & \textbf{Hj} & \textbf{Lzh} &                &                &                &                \\ \midrule
      \ding{52}                                    &             &              & 25.96          & 8.02           & 4.50           & 10.29          \\
      & \ding{52}   &              & 13.82          & {\ul 48.97}    & 33.07          & 12.32          \\
      &             & \ding{52}    & 19.08          & 9.79           & 4.85           & 18.13          \\
      \ding{52}                                    & \ding{52}   &              & \textbf{45.13} & \textbf{49.53} & \textbf{34.69} & 14.00          \\
      \ding{52}                                    &             & \ding{52}    & 37.10          & 9.70           & 4.85           & 17.88          \\
      & \ding{52}   & \ding{52}    & 19.14          & 47.49          & {\ul 33.91}    & \textbf{18.78} \\
      \ding{52}                                    & \ding{52}   & \ding{52}    & {\ul 42.66}    & 47.93          & 33.69          & {\ul 18.40}    \\ \bottomrule
    \end{tabular}%
  }
  \caption{Translation performance (BLEU score) comparison across different combinations of Kanbun (Kb, 0.34M tokens), Hanja (351.1M tokens), and Classical Chinese (79.8M tokens) training data. The \textbf{bold} and \underline{underlined} values indicate the best and second-best performance, respectively.}
  \label{tab:kanbun-mt}
\end{table}

%% file: figures/vocab.tex
\begin{figure}[t!]
  \centering
  \includegraphics[width=\columnwidth]{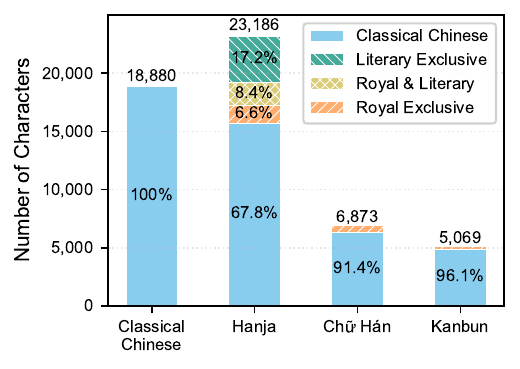}
  \caption{Distribution of unique characters across writing systems in the Sinosphere. The bars represent the proportion of shared characters with Classical Chinese versus language-specific variants in each writing system.}
  \label{fig:vocab}
\end{figure}

%% file: figures/vocab-heatmap.tex
\begin{figure}[t!]
  \centering
  \includegraphics[width=\columnwidth]{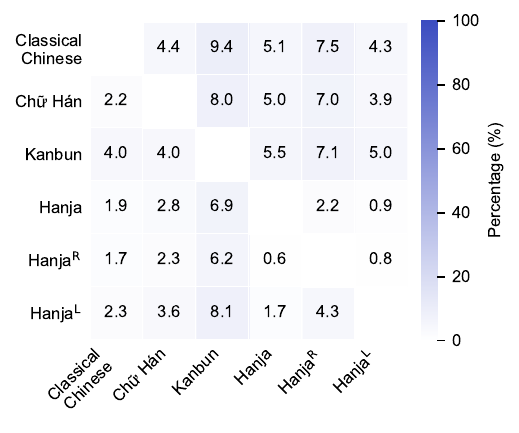}
  \caption{Heatmap of character coverage gaps between Sinosphere languages. Each cell shows the percentage of characters in the \emph{row} language that are not the most common characters in the \emph{column} language at 99\% frequency threshold.}
  \label{fig:vocab-heatmap}
\end{figure}

%% file: contents/5_related_work.tex
\subsection{NLP for Asian Historical Documents}

A variety of research has been mainly conducted in Classical Chinese and Hanja due to challenges for acquisition of available resources.
In Classical Chinese, evaluation datasets and benchmarks~\citep{zhou-etal-2023-wyweb} and language models~\citep{tian-2021-anchibert, chang-2023-sikugpt} are widely released.
Similarly, datasets and language models for Hanja have been introduced for various tasks, including machine translation~\citep{kang-etal-2021-restoring, son-etal-2022-translating}, named entity recognition~\citep{yoo-etal-2022-hue}, and relation extraction~\citep{yang-etal-2023-histred}.

\subsection{Cross-Lingual Studies for Sinosphere}

Several studies have introduced cross-lingual approaches that leverage linguistically close, historical resources in the Sinosphere.
\citet{moon-2024-exploiting} used Classical Chinese resources to develop NER and sentence splitting models for Hanja literary documents and uncovered that removing special characters and punctuation marks helps cross-lingual transfer between Classical Chinese and Hanja.
\citet{wang-etal-2023-kanbun} synthetically constructed the first Classical Chinese-to-Kanbun dataset and trained a Kanbun language model, addressing the scarcity of available resources in Kanbun.

Cross-lingual transfer in the Sinosphere has also been explored across modern languages.
\citet{kim-etal-2020-korean} proposed a machine translation technique that matches overlapping vocabulary between Korean and Japanese stemming from Hanja and Kanbun, respectively.
\citet{nehrdich-etal-2023-mitra} used Classical Chinese-to-Modern Chinese dataset for Buddhist Chinese-to-English machine translation.
While recent studies have recklessly adopted Classical Chinese resources for other languages in the Sinosphere, this paper aims to carefully investigate the performance of cross-lingual transfer.

%% file: contents/6_conclusion.tex
We challenge the widespread assumption that Classical Chinese resources inherently benefit language models for other historical East Asian writing systems.
Our comprehensive experiments across machine translation, named entity recognition, and punctuation restoration reveal that incorporating Classical Chinese data produces minimal and often statistically insignificant improvements for Hanja documents.
While our analysis shows limited character-level divergence between these languages, the poor cross-lingual transfer suggests fundamental linguistic differences beyond shared vocabulary.
These findings demonstrate that successfully processing historical Asian languages requires careful empirical validation rather than assumed benefits from apparent linguistic similarities.
We emphasize the importance of considering both resource availability and domain characteristics when developing language models for historical documents.
Building on our results, future research should further investigate the linguistic factors that affect cross-lingual transferability across different languages or writing systems.

%% file: contents/7_limitation.tex
Our experiments with Kanbun and \chuhan{} are constrained by limited dataset availability compared to Hanja, necessitating caution in drawing broader conclusions about these writing systems.
Also, as NLP researchers rather than domain experts in historical Asian languages, our analysis may not fully capture deeper linguistic nuances in ancient languages.

Despite analyzing substantial volumes of historical records and literary work, our coverage of Hanja documents remains partial.
Notable omissions include local government records, Buddhist texts, and epigraphic sources, which may demonstrate distinct patterns of cross-lingual transferability from Classical Chinese.

The representation of Classical Chinese texts in our datasets poses an additional limitation, as they are available only in Simplified Chinese despite their Traditional Chinese origins.
This inherently imperfect character conversion system may introduce systematic biases in our cross-lingual analysis.

%% file: contents/8_ethics.tex
This research focuses on evaluating the effectiveness of cross-lingual transfer between historical writing systems through computational experiments on publicly available historical documents.
The methods employed are applied to texts that have been openly preserved for academic study.
The research does not involve human subjects, sensitive personal data, or content that could enable harmful applications.
While historical texts can sometimes contain biased perspectives or sensitive content, our work focuses purely on the technical aspects of language processing rather than interpreting or generating content.
The computational methods and findings presented here aim to advance the scholarly study of historical documents while maintaining respect for the cultural significance of these texts.

%% file: contents/9_ack.tex
This work was supported by Institute of Information \& communications Technology Planning \& Evaluation (IITP) grant funded by the Korea government (MSIT) (No. RS-2024-00509258 and No. RS-2024-00469482, Global AI Frontier Lab).

This work was also supported by the Samsung Advanced Institute of Technology (under the project Next Generation Deep Learning: From Pattern Recognition to AI).

This research project has benefitted from the Microsoft Accelerate Foundation Models Research (AFMR) grant program through which leading foundation models hosted by Microsoft Azure along with access to Azure credits were provided to conduct the research.

We acknowledge using Claude\footnote{\ztexttt{\url{https://claude.ai}}} and   ChatGPT\footnote{\ztexttt{\url{https://chatgpt.com}}} for writing and coding assistance.

%% file: contents/appendix.tex
\section{Replication Details}
\label{sec:appx_replication_details}

\subsection{Data Sources}
\label{sec:appx_data_sources}

We collect our datasets from publicly available sources between February and October 2024.
Korean historical documents are sourced from national research institutions: the National Institute of Korean History (NIKH) provides the AJD\footnote{\ztexttt{\url{https://sillok.history.go.kr}}} and DRS\footnote{\ztexttt{\url{https://sjw.history.go.kr}}}, while the Kyujanggak Institute maintains DRRI\footnote{\ztexttt{\url{https://kyudb.snu.ac.kr/series/main.do?item_cd=ILS}}}.
The Institute for the Translation of Korean Classics (ITKC) offers the KLC\footnote{\ztexttt{\url{https://db.itkc.or.kr}}} along with Korean translations of the royal documents.
Classical Chinese resources include Daizhige\footnote{\ztexttt{\url{https://github.com/garychowcmu/daizhigev20}}}, NiuTrans\footnote{\ztexttt{\url{https://github.com/NiuTrans/Classical-Modern}}}, C2MChn\footnote{\ztexttt{\url{https://github.com/Zongyuan-Jiang/C2MChn}}}, and WYWEB\footnote{\ztexttt{\url{https://github.com/baudzhou/WYWEB}}}, all available through GitHub repositories.
The OCDB\footnote{\ztexttt{\url{https://db.cyberseodang.or.kr}}} is maintained by the Institute of Traditional Culture.
For Japanese documents, we use the Rikkokushi texts from the public website\footnote{\ztexttt{\url{http://www.kikuchi2.com/sheet/rikkoku.html}}}, with Korean translations of Korea-related records provided by NIKH\footnote{\ztexttt{\url{https://db.history.go.kr/id/jm}}}.
Vietnamese historical chronicles including ĐVSKTT\footnote{\ztexttt{\url{https://zh.wikisource.org/wiki/}}\btexttt{大越史記全書}}, ĐNTL\footnote{\ztexttt{\url{https://zh.wikisource.org/wiki/}}\btexttt{大南寔錄}}, ANCL\footnote{\ztexttt{\url{https://zh.wikisource.org/wiki/}}\btexttt{安南志畧}}, and ĐVSL\footnote{\ztexttt{\url{https://zh.wikisource.org/wiki/}}\btexttt{越史略}} are available through Wikisource.

\subsection{Data Augmentation}
\label{sec:appx_data_aug}

We create synthetic Korean translations of Classical Chinese texts using GPT-4.
For each source text, we provide both the Classical Chinese original and its Modern Chinese translation as context, using the following prompt:
\begin{tcolorbox}[breakable, enhanced, top=1pt, left=1pt, right=1pt, bottom=1pt]
  \small{
    Translate the following text from Classical Chinese into Korean, based on the reference translation in Modern Chinese. \keys{\return} \\
    Classical Chinese: <source sentence> \keys{\return} \\
    Modern Chinese: <reference translation> \keys{\return} \\
    Korean:
  }
\end{tcolorbox}
We generate translations using GPT-4 under two configurations: the NiuTrans dataset translations use \ztexttt{gpt-4-0125-preview} with temperature 0.7, while C2MChn translations use \ztexttt{gpt-4o-mini-2024-07-18} with temperature 0.0.
We employ Azure OpenAI Service as our primary platform, falling back to the OpenAI API when necessary.
Approximately 6\% of source texts are filtered out due to sensitive historical content, particularly passages containing references to war crimes or violence.

\subsection{Preprocessing}
\label{sec:appx_preprocessing}

Processing ancient Asian texts requires careful character normalization to ensure consistent representation across different writing systems and time periods.
Our preprocessing pipeline applies the Normalization Form Compatibility Composition (NFKC) to standardize character encodings, followed by whitespace standardization that converts all newlines, tabs, and spaces to single space characters.
We normalize all punctuation marks, including converting directional quotation marks to their neutral forms, and standardize CJK middle dot variants (\ztexttt{U+318D}, \ztexttt{U+119E}, \ztexttt{U+30FB}) to the standard middle dot form (\ztexttt{U+00B7}).
For Classical Chinese texts in Simplified Chinese characters, we convert them to Traditional Chinese using OpenCC\footnote{\ztexttt{\url{https://github.com/BYVoid/OpenCC}}}.

\subsection{Experimental Setup}
\label{sec:appx_exp_setup}

\input{tables/train_data_composition}
\input{tables/task_data_size}

Table~\ref{tab:train_data_composition} quantifies our experimental data across tasks using both sample counts and token quantities.
Table~\ref{tab:task_data_size} presents our dataset partitioning across training, validation, and test sets for each task.
For machine translation (MT), we evaluate performance using 1,000 test samples per document and language pair, computing aggregate BLEU scores via SacreBLEU across all translation outputs.
For named entity recognition (NER) and punctuation restoration (PR), we use 5,000 test samples per document, with the exception of GLNER, which uses 2,000 test samples due to dataset constraints.

\subsection{Training and Hyperparameters}
\label{sec:appx_training_hparams}

\input{tables/hparam}

Our experiments run on a server equipped with Intel Xeon Silver 4114 processor (40 threads) and eight GeForce RTX 2080 Ti GPUs (11GB each).
For training and inference of Gemma-2 models, we use a separate server with Intel Xeon Silver 4214R processor (48 threads) and eight Quadro RTX A6000 GPUs (48GB each).
We implement our models using LLaMA-Factory~\citep{zheng-etal-2024-llamafactory} for machine translation fine-tuning and Hugging Face Transformers~\citep{wolf-etal-2020-transformers} for NER and PR models.
Table~\ref{tab:hparam} details our hyperparameter configurations.
Training times vary by task: up to 36 hours for machine translation, 10 hours for named entity recognition, and 14 hours for punctuation restoration.
The prompt shown below is used consistently across all translation tasks during both training and inference.

\begin{tcolorbox}[breakable, enhanced, top=1pt, left=1pt, right=1pt, bottom=1pt]
  \small{
    Translate the following text from <source language> into <target language>. \keys{\return}

    <source language>: <source sentence> \keys{\return}

    <target language>:
  }
\end{tcolorbox}

\input{tables/klc_mt_genre}

\subsection{Inference and Evaluation}
\label{sec:appx_inference_evaluation}

\input{tables/metric}
\input{tables/punc_reduction}

\paragraph{Machine Translation.}
We quantize the fine-tuned MT models using AWQ~\citep{lin-2023-awq} and utilize vLLM~\citep{kwon-2023-efficient} for inference.
The prompt used for training is also used for inference.
We set the temperature to 0 and employ greedy decoding.
Metric signatures and versions used for evaluation are presented in Table~\ref{tab:metric}.

\paragraph{Punctuation Restoration.}
For evaluation, we simplify the diverse punctuation marks used in the original documents and our models into a standardized 4-class scheme consisting of COMMA, PERIOD, QUESTION, and OTHER.
This allows for consistent comparison of model performance across the different datasets.
Table~\ref{tab:punc_reduction} shows how various punctuation characters are mapped to these four classes based on their typical functions or meanings.

\subsection{Korean Literary Collections Dataset}
\label{sec:appx_klc_dataset}

For this study, we compile a new dataset from the Korean Literary Collections (KLC), a comprehensive collection of Hanja literary works maintained by the Institute for the Translation of Korean Classics.
Unlike prior research that focused predominantly on royal-centric Hanja documents \citep{kang-etal-2021-restoring, yoo-etal-2022-hue, son-etal-2022-translating}, our KLC dataset captures diverse writing styles from individual scholars spanning from 886 to 1933 CE, with particularly rich coverage during the 1800s-1930s period.
The source corpus contains 652,622 unique articles with an average length of 337 Hanja characters (approximately 220M characters total) from 1,258 unique authors, including notable historical figures such as Song Si-yeol (宋時烈), Jeong Yak-yong (丁若鏞), and Kwak Jong-seok (郭鍾錫).
Table~\ref{tab:klc_mt_genre} presents the genre distribution of the translated portion, demonstrating substantial coverage beyond official documents.
We structured our KLC dataset to support multiple NLP tasks: raw text for language model pretraining (652,622 samples), parallel data for machine translation (157,202 samples with Hanja-Korean translations), and annotated documents for named entity recognition (21,657 samples with 379,976 entities).

\section{Complementary Results}
\label{sec:appx_complementary_results}

This section presents additional experimental results and analyses that complement our main findings.

\subsection{Experimental Results}
\label{sec:appx_exp_results}

\input{tables/all_bleu_score}

Table~\ref{tab:all_bleu_score} provides comprehensive BLEU scores for machine translation experiments across all dataset combinations and language pairs, including results from different model architectures and training configurations.

\subsection{Threshold for Diminishing Benefits}
\label{sec:appx_threshold_diminishing}

\input{tables/low-resource_all}

Table~\ref{tab:low-resource_all} details our systematic investigation of how varying the ratio between Hanja and Classical Chinese training data affects model performance.
The results encompass performance metrics across machine translation, named entity recognition, and punctuation restoration tasks as we gradually reduce the proportion of Hanja data on a logarithmic scale.

\subsection{Machine Translation for Kanbun}
\label{sec:appx_mt_kanbun}

\input{figures/kanbun-low}

Figure~\ref{fig:kanbun-low} illustrates how BLEU scores change as the quantity of additional training data decreases for Kanbun-Korean translation.
The relative performance advantages between different systems remain consistent across varying data quantities.

\subsection{Vocabulary Divergence}
\label{sec:appx_vocab_divergence}

\input{figures/vocab-heatmap-all}

Figure~\ref{fig:vocab-heatmap-all} presents the proportion of unique characters in each corpus that do not appear in other corpora, measured at four cumulative frequency thresholds: 100\%, 99.9\%, 99\%, and 95\%.
This analysis reveals the extent of character-level divergence between writing systems in the Sinosphere.

\subsection{Analysis of Performance in Low-Resource Settings}
\label{sec:appx_analysis_low}

To verify that the benefits observed when adding Classical Chinese resources in low-resource scenarios reflect genuine cross-lingual transfer rather than overfitting mitigation, we conduct additional experiments analyzing evaluation loss behavior.
We maintain full validation set sizes while systematically reducing training data for Hanja-only models at two extreme low-resource ratios (1/16 and 1/32 of the original data).
Figure~\ref{fig:low_resource_loss} shows that evaluation loss decreases monotonically across all settings for both NER and PR tasks, with no indication of validation loss increases or plateauing that would typically signal overfitting.
This consistent pattern across different data ratios strongly suggests that models trained on extremely limited Hanja data do not suffer from overfitting, even without Classical Chinese data.
Therefore, the performance improvements observed when adding Classical Chinese resources in these settings likely represent genuine benefits from cross-lingual transfer rather than simply regularization effects addressing overfitting issues.

\input{figures/low_resource_loss}

\subsection{Qualitative Error Analysis}
\label{sec:appx_error_analysis}

To complement our quantitative findings, we perform a systematic qualitative analysis comparing outputs of models trained with and without Classical Chinese data.
We calculate per-sample performance metrics for all test predictions and categorize instances where the inclusion of Classical Chinese resources leads to performance changes.
Table~\ref{tab:error_analysis} presents representative examples from our analysis of Hanja$^{\text{R}}$ to Korean translation, which reveals three recurring error patterns: (1)~inappropriate modernization of classical terms, where historically specific terminology is simplified into contemporary equivalents (\eg ``찬구(饌具)'' $\rightarrow$ ``반찬'', replacing a formal historical term for food provisions with a modern casual word for side dishes); (2)~loss of Korea-specific concepts, where terms unique to Korean historical and cultural contexts are omitted or generalized (\eg ``황의장(黃儀仗)'' $\rightarrow$ ``의장'', losing the Korea-specific royal ceremonial context); and (3)~name translation errors, where historical Korean names are inconsistently handled (\eg ``윤방(尹滂)'' $\rightarrow$ ``윤팽'', incorrectly changing the pronunciation).
These patterns suggest that Classical Chinese data can introduce biases that obscure culturally and historically specific nuances in Hanja translation, explaining the quantitative performance degradation observed in \cref{sec:exp_results}.
For sequence labeling tasks (NER and PR), our analysis shows no consistent patterns of improvement or degradation, aligning with the statistical non-significance reported in our main results.

\input{tables/error_analysis}

%% file: tables/train_data_composition.tex
\begin{table}[b!]
  \centering
  \resizebox{\columnwidth}{!}{%
    \begin{tabular}{@{}lllrr@{}}
      \toprule
      \textbf{Task}          &
      \textbf{Type}          &
      \textbf{Document}      &
      \textbf{\# of Samples} &
      \textbf{\# of Tokens}                                                       \\ \midrule
      \multirow{3}{*}{MT}    & Hj$^{\text{R}}$ & AJD      & 331,150 & 241,653,871 \\
      & Hj$^{\text{L}}$ & KLC      & 53,147  & 109,406,346 \\
      & Lzh             & NiuTrans & 774,914 & 79,806,362  \\ \midrule
      \multirow{3}{*}{NER}   & Hj$^{\text{R}}$ & AJD      & 293,854 & 80,841,316  \\
      & Hj$^{\text{L}}$ & KLC      & 8,035   & 6,673,763   \\
      & Lzh             & GLNER    & 14,719  & 4,710,310   \\ \midrule
      \multirow{3}{*}{PR}    & Hj$^{\text{R}}$ & AJD      & 293,746 & 81,095,372  \\
      & Hj$^{\text{L}}$ & KLC      & 14,428  & 7,983,038   \\
      & Lzh             & WYWEB    & 70,664  & 13,141,862  \\ \bottomrule
    \end{tabular}%
  }
  \caption{Composition of training data used in experiments across tasks. Data quantities are shown by both number of samples and total tokens computed using \ztexttt{cl100k\_base} encoding.}
  \label{tab:train_data_composition}
\end{table}

%% file: tables/task_data_size.tex
\begin{table}[t!]
  \centering
  \resizebox{\columnwidth}{!}{%
    \begin{tabular}{@{}lllcrrr@{}}
      \toprule
      \textbf{Tasks}       & \textbf{Type}                    & \textbf{Document}    & \textbf{Lang.} & \textbf{Train} & \textbf{Val} & \textbf{Test} \\ \midrule
      \multirow{11}{*}{MT} & \multirow{5}{*}{Hj$^{\text{R}}$} & \multirow{3}{*}{AJD} & Hj-En          & 16,032         & 0            & 1,000         \\
      &                                  &                      & Hj-Ko          & 299,106        & 0            & 1,000         \\
      &                                  &                      & Ko-En          & 16,012         & 0            & 1,000         \\
      &                                  & DRS                  & Hj-Ko          & 0              & 0            & 1,000         \\
      &                                  & DRRI                 & Hj-Ko          & 0              & 0            & 1,000         \\
      & Hj$^{\text{L}}$                  & KLC                  & Hj-Ko          & 53,147         & 0            & 1,000         \\
      & \multirow{4}{*}{Lzh}             & NiuTrans             & Lzh-Ko         & 774,914        & 0            & 1,000         \\
      &                                  & WYWMT                & Lzh-Ko         & 0              & 0            & 1,000         \\
      &                                  & OCDB                 & Lzh-Ko         & 0              & 0            & 1,000         \\
      &                                  & C2MChn$^\dagger$     & Lzh-Ko         & 542,305        & 0            & 0             \\
      & Kb                               & Rikkokushi$^\dagger$ & Kb-Ko          & 1,025          & 0            & 346           \\ \midrule
      \multirow{3}{*}{NER} & Hj$^{\text{R}}$                  & AJD                  & Hj             & 293,854        & 37,830       & 5,000         \\
      & Hj$^{\text{L}}$                  & KLC                  & Hj             & 8,035          & 995          & 5,000         \\
      & Lzh                              & GLNER                & Lzh            & 14,719         & 2,000        & 2,000         \\ \midrule
      \multirow{3}{*}{PR}  & Hj$^{\text{R}}$                  & AJD                  & Hj             & 293,746        & 37,831       & 5,000         \\
      & Hj$^{\text{L}}$                  & KLC                  & Hj             & 14,428         & 1,797        & 5,000         \\
      & Lzh                              & WYWEB                & Lzh            & 70,664         & 32,607       & 5,000         \\ \bottomrule
    \end{tabular}%
  }
  \caption{Dataset composition and partitioning across tasks. The table shows sample sizes for training, validation, and test sets used in machine translation (MT), named entity recognition (NER), and punctuation restoration (PR) experiments. Documents marked with $\dagger$ are supplementary materials used only in discussions.}
  \label{tab:task_data_size}
\end{table}

%% file: tables/hparam.tex
\begin{table}[t!]
  \centering

  \begin{subtable}{\linewidth}
    \input{tables/hparam_mt}
  \end{subtable}
  \vspace{1mm}

  \begin{subtable}{\linewidth}
    \input{tables/hparam_ner_pr}
  \end{subtable}

  \caption{Hyperparameter configurations for training MT, NER, and PR models. Values shown for MT models use \ztexttt{Qwen/Qwen2-7B} base architecture (additional experiments use \ztexttt{Qwen/Qwen2-1.5B}, \ztexttt{Qwen/Qwen2-0.5B}, \ztexttt{google/gemma-2-9b}, and \ztexttt{meta-llama/Llama-3.1-8B-Instruct}). We use half precision (\ztexttt{fp16}) for all computation.}
  \label{tab:hparam}
\end{table}

%% file: tables/hparam_mt.tex
\centering
\resizebox{\linewidth}{!}{%
  \begin{tabular}{@{}ll@{}}
    \toprule
    \textbf{Hyperparameter} & \textbf{Value}                               \\ \midrule
    Max sequence length     & 512                                          \\
    Batch size              & 64                                           \\
    Initial checkpoint      & \ztexttt{Qwen/Qwen2-7B}                      \\
    Quantization            &
    \begin{tabular}[c]{@{}l@{}}4-bit NormalFloat \\ and double quantization
    \end{tabular} \\
    LoRA \textit{r}         & 16                                           \\
    LoRA $\alpha$           & 32                                           \\
    LoRA dropout            & 0.0                                          \\
    rsLoRA                  & True                                         \\
    Number of epochs        & 1                                            \\
    Learning rate           & 1.0e-4                                       \\
    Learning rate scheduler & Cosine                                       \\
    Warm-up ratio           & 0.1                                          \\
    Optimizer               & 8-bit AdamW                                  \\
    Weight decay            & 0.01                                         \\
    Gradient clipping       & 1.0                                          \\ \bottomrule
  \end{tabular}%
}
\caption{Hyperparameters for MT models.}
\label{tab:hparam_mt}

%% file: tables/hparam_ner_pr.tex
\centering
\resizebox{\linewidth}{!}{%
  \begin{tabular}{@{}ll@{}}
    \toprule
    \textbf{Hyperparameter} & \textbf{Value}                  \\ \midrule
    Max sequence length     & 512                             \\
    Batch size              & 32                              \\
    Initial checkpoint      & \ztexttt{SIKU-BERT/sikuroberta} \\
    Max epochs              & 5                               \\
    Early stopping          & applied on validation loss      \\
    Learning rate           & 2e-4                            \\
    Learning rate scheduler & Linear                          \\
    Warm-up ratio           & 0.1                             \\
    Optimizer               & AdamW                           \\
    Weight decay            & 0.01                            \\ \bottomrule
  \end{tabular}%
}
\caption{Hyperparameters for NER and PR models.}
\label{tab:hparam_ner_pr}

%% file: tables/klc_mt_genre.tex
\begin{table}[b!]
  \centering
  \begin{tabular}{@{}lrr@{}}
    \toprule
    \textbf{Genre}      & \textbf{\# of Articles} & \textbf{Ratio (\%)} \\ \midrule
    Anthology           & 112,215                 & 71.4                \\
    Miscellaneous       & 11,707                  & 7.4                 \\
    Travelogue          & 9,688                   & 6.2                 \\
    Literature          & 5,456                   & 3.5                 \\
    History             & 4,035                   & 2.6                 \\
    Complete Collection & 3,878                   & 2.5                 \\
    Law                 & 1,609                   & 1.0                 \\
    Ceremonial Texts    & 1,593                   & 1.0                 \\
    Human Affairs       & 1,422                   & 0.9                 \\
    Astronomy           & 1,075                   & 0.7                 \\
    Politics            & 944                     & 0.6                 \\
    Medicine            & 779                     & 0.5                 \\
    Geography           & 664                     & 0.4                 \\
    Agriculture         & 618                     & 0.4                 \\
    Philosophy          & 595                     & 0.4                 \\
    Ceremonial Records  & 452                     & 0.3                 \\
    Foreign Relations   & 364                     & 0.2                 \\
    Classical Texts     & 53                      & 0.0                 \\
    Mathematics         & 41                      & 0.0                 \\
    Archives            & 14                      & 0.0                 \\ \bottomrule
  \end{tabular}
  \caption{Genre distribution of the translated portion in the Korean Literary Collections (KLC) dataset, showing the number of articles and percentage for each category.}
  \label{tab:klc_mt_genre}
\end{table}

%% file: tables/metric.tex
\begin{table*}[b!]
  \centering
  \begin{tabular}{@{}ll@{}}
    \toprule
    \textbf{Metric}                                                          & \textbf{Version}                                                         \\ \midrule
    BLEU {[}En{]}                                                            & \ztexttt{nrefs:1|case:mixed|eff:no|tok:13a|smooth:exp|version:2.4.2}     \\
    \begin{tabular}[c]{@{}l@{}}BLEU {[}En{]} Paired- \\ bootstrap resampling
    \end{tabular} &
    \begin{tabular}[c]{@{}l@{}}\ztexttt{nrefs:1|bs:2000|seed:42|case:mixed|eff:no|tok:13a|smooth:exp|} \\ \ztexttt{version:2.4.2}
    \end{tabular}                        \\
    BLEU {[}Ko{]}                                                            &
    \begin{tabular}[c]{@{}l@{}}\ztexttt{nrefs:1|case:mixed|eff:no|tok:ko-mecab-0.996/ko-0.9.2-KO|} \\ \ztexttt{smooth:exp|version:2.4.2}
    \end{tabular}                 \\
    \begin{tabular}[c]{@{}l@{}}BLEU {[}Ko{]} Paired- \\ bootstrap resampling
    \end{tabular} &
    \begin{tabular}[c]{@{}l@{}}\ztexttt{nrefs:1|bs:2000|seed:42|case:mixed|eff:no|tok:ko-mecab-0.996/} \\ \ztexttt{ko-0.9.2-KO|smooth:exp|version:2.4.2}
    \end{tabular} \\
    BLEU {[}Zh{]}                                                            & \ztexttt{nrefs:1|case:mixed|eff:no|tok:zh|smooth:exp|version:2.4.2}      \\
    \begin{tabular}[c]{@{}l@{}}BLEU {[}Zh{]} Paired- \\ bootstrap resampling
    \end{tabular} &
    \begin{tabular}[c]{@{}l@{}}\ztexttt{nrefs:1|bs:2000|seed:42|case:mixed|eff:no|tok:zh|smooth:exp|} \\ \ztexttt{version:2.4.2}
    \end{tabular}                         \\
    \bottomrule
  \end{tabular}%
  \caption{Metric versions and signatures.}
  \label{tab:metric}
\end{table*}

%% file: tables/punc_reduction.tex
\begin{table*}[htb!]
  \centering
  \begin{tabular}{@{}ll@{}}
    \toprule
    \textbf{Class} & \textbf{Characters}                                                              \\ \midrule
    COMMA          & \ztexttt{- (U+002D), / (U+002F), : (U+003A), | (U+007C), · (U+00B7), 、 (U+3001)} \\
    PERIOD         & \ztexttt{! (U+0021), . (U+002E), ; (U+003B), 。 (U+3002)}                         \\
    QUESTION       & \ztexttt{? (U+003F)}                                                             \\ \bottomrule
  \end{tabular}%
  \caption{Punctuation reduction rules for simplifying diverse punctuation marks in the punctuation restoration task to a standardized 4-class scheme: COMMA, PERIOD, QUESTION, and OTHER.}
  \label{tab:punc_reduction}
\end{table*}

%% file: tables/all_bleu_score.tex
\begin{sidewaystable*}[htb!]
  \centering
  \resizebox{\textwidth}{!}{%
    \begin{tabular}{@{}c|ccccccc|ccccccccccc@{}}
      \toprule
      \multirow{4}{*}{\textbf{Model}}                      &
      \multicolumn{7}{c|}{\textbf{Train Data}}             &
      \multicolumn{11}{c}{\textbf{Test Data (BLEU)}}         \\
      &
      \multicolumn{1}{c|}{\textbf{Hj$^\text{R}$}}          &
      \multicolumn{1}{c|}{\textbf{Hj$^\text{L}$}}          &
      \multicolumn{4}{c|}{\textbf{Lzh}}                    &
      \textbf{Kb}                                          &
      \multicolumn{4}{c|}{\textbf{Hj$^\text{R}$}}          &
      \multicolumn{1}{c|}{\textbf{Hj$^\text{L}$}}          &
      \multicolumn{5}{c|}{\textbf{Lzh}}                    &
      \textbf{Kb}                                            \\
      &
      \multicolumn{1}{c|}{\multirow{2}{*}{\textbf{AJD}}}   &
      \multicolumn{1}{c|}{\multirow{2}{*}{\textbf{KLC}}}   &
      \multirow{2}{*}{\textbf{
          \begin{tabular}[c]{@{}c@{}}Niu \\ Trans
      \end{tabular}}}              &
      \multicolumn{3}{c|}{\textbf{C2MChn}}                 &
      \multirow{2}{*}{\textbf{
          \begin{tabular}[c]{@{}c@{}}Rikko- \\ kushi
      \end{tabular}}}           &
      \multicolumn{2}{c}{\textbf{AJD}}                     &
      \textbf{DRS}                                         &
      \multicolumn{1}{c|}{\textbf{DRRI}}                   &
      \multicolumn{1}{c|}{\textbf{KLC}}                    &
      \textbf{OCDB}                                        &
      \multicolumn{2}{c}{\textbf{NiuTrans}}                &
      \multicolumn{2}{c|}{\textbf{WYWMT}}                  &
      \textbf{Rikkokushi}                                    \\
      &
      \multicolumn{1}{c|}{}                                &
      \multicolumn{1}{c|}{}                                &
      &
      \textbf{His}                                         &
      \textbf{Rel}                                         &
      \multicolumn{1}{c|}{\textbf{Mis}}                    &
      &
      \textbf{Hj-En}                                       &
      \textbf{Hj-Ko}                                       &
      \textbf{Hj-Ko}                                       &
      \multicolumn{1}{c|}{\textbf{Hj-Ko}}                  &
      \multicolumn{1}{c|}{\textbf{Hj-Ko}}                  &
      \textbf{Lzh-Ko}                                      &
      \textbf{Lzh-Ko}                                      &
      \textbf{Lzh-Zh}                                      &
      \textbf{Lzh-Ko}                                      &
      \multicolumn{1}{c|}{\textbf{Lzh-Zh}}                 &
      \textbf{Kb-Ko}                                         \\ \midrule
      \multirow{7}{*}{Qwen2-7B}                            &
      \multicolumn{1}{c|}{-}                               &
      \multicolumn{1}{c|}{-}                               &
      \ding{52}                                            &
      -                                                    &
      -                                                    &
      \multicolumn{1}{c|}{-}                               &
      -                                                    &
      0.02                                                 &
      10.96                                                &
      10.35                                                &
      \multicolumn{1}{c|}{7.22}                            &
      \multicolumn{1}{c|}{4.85}                            &
      12.93                                                &
      26.25                                                &
      5.75                                                 &
      21.60                                                &
      \multicolumn{1}{c|}{6.18}                            &
      19.08                                                  \\
      &
      \multicolumn{1}{c|}{\ding{52}}                       &
      \multicolumn{1}{c|}{-}                               &
      -                                                    &
      -                                                    &
      -                                                    &
      \multicolumn{1}{c|}{-}                               &
      -                                                    &
      33.16                                                &
      55.13                                                &
      47.39                                                &
      \multicolumn{1}{c|}{39.64}                           &
      \multicolumn{1}{c|}{10.81}                           &
      14.63                                                &
      9.13                                                 &
      20.70                                                &
      7.26                                                 &
      \multicolumn{1}{c|}{13.38}                           &
      -                                                      \\
      &
      \multicolumn{1}{c|}{\ding{52}}                       &
      \multicolumn{1}{c|}{-}                               &
      \ding{52}                                            &
      -                                                    &
      -                                                    &
      \multicolumn{1}{c|}{-}                               &
      -                                                    &
      31.34                                                &
      52.49                                                &
      46.40                                                &
      \multicolumn{1}{c|}{39.03}                           &
      \multicolumn{1}{c|}{11.82}                           &
      13.71                                                &
      26.65                                                &
      18.58                                                &
      21.62                                                &
      \multicolumn{1}{c|}{14.02}                           &
      -                                                      \\
      &
      \multicolumn{1}{c|}{-}                               &
      \multicolumn{1}{c|}{\ding{52}}                       &
      -                                                    &
      -                                                    &
      -                                                    &
      \multicolumn{1}{c|}{-}                               &
      -                                                    &
      0.13                                                 &
      38.34                                                &
      34.67                                                &
      \multicolumn{1}{c|}{28.22}                           &
      \multicolumn{1}{c|}{33.57}                           &
      14.11                                                &
      9.88                                                 &
      20.22                                                &
      8.53                                                 &
      \multicolumn{1}{c|}{10.73}                           &
      -                                                      \\
      &
      \multicolumn{1}{c|}{-}                               &
      \multicolumn{1}{c|}{\ding{52}}                       &
      \ding{52}                                            &
      -                                                    &
      -                                                    &
      \multicolumn{1}{c|}{-}                               &
      -                                                    &
      0.06                                                 &
      35.59                                                &
      30.22                                                &
      \multicolumn{1}{c|}{26.11}                           &
      \multicolumn{1}{c|}{32.19}                           &
      12.94                                                &
      26.12                                                &
      10.51                                                &
      21.57                                                &
      \multicolumn{1}{c|}{8.66}                            &
      -                                                      \\
      &
      \multicolumn{1}{c|}{\ding{52}}                       &
      \multicolumn{1}{c|}{\ding{52}}                       &
      -                                                    &
      -                                                    &
      -                                                    &
      \multicolumn{1}{c|}{-}                               &
      -                                                    &
      33.15                                                &
      55.30                                                &
      48.65                                                &
      \multicolumn{1}{c|}{40.65}                           &
      \multicolumn{1}{c|}{33.07}                           &
      16.13                                                &
      9.42                                                 &
      15.13                                                &
      7.33                                                 &
      \multicolumn{1}{c|}{8.74}                            &
      13.82                                                  \\
      &
      \multicolumn{1}{c|}{\ding{52}}                       &
      \multicolumn{1}{c|}{\ding{52}}                       &
      \ding{52}                                            &
      -                                                    &
      -                                                    &
      \multicolumn{1}{c|}{-}                               &
      -                                                    &
      31.52                                                &
      52.83                                                &
      47.04                                                &
      \multicolumn{1}{c|}{39.33}                           &
      \multicolumn{1}{c|}{33.91}                           &
      14.26                                                &
      26.06                                                &
      1.21                                                 &
      21.68                                                &
      \multicolumn{1}{c|}{0.86}                            &
      19.14                                                  \\ \midrule
      \multirow{2}{*}{Qwen2-1.5B}                          &
      \multicolumn{1}{c|}{\ding{52}}                       &
      \multicolumn{1}{c|}{\ding{52}}                       &
      -                                                    &
      -                                                    &
      -                                                    &
      \multicolumn{1}{c|}{-}                               &
      -                                                    &
      28.74                                                &
      50.69                                                &
      43.32                                                &
      \multicolumn{1}{c|}{35.02}                           &
      \multicolumn{1}{c|}{29.32}                           &
      11.12                                                &
      7.66                                                 &
      1.78                                                 &
      5.42                                                 &
      \multicolumn{1}{c|}{0.92}                            &
      -                                                      \\
      &
      \multicolumn{1}{c|}{\ding{52}}                       &
      \multicolumn{1}{c|}{\ding{52}}                       &
      \ding{52}                                            &
      -                                                    &
      -                                                    &
      \multicolumn{1}{c|}{-}                               &
      -                                                    &
      23.66                                                &
      45.58                                                &
      36.02                                                &
      \multicolumn{1}{c|}{29.89}                           &
      \multicolumn{1}{c|}{26.66}                           &
      11.03                                                &
      23.14                                                &
      0.11                                                 &
      18.30                                                &
      \multicolumn{1}{c|}{0.05}                            &
      -                                                      \\ \midrule
      \multirow{2}{*}{Qwen2-0.5B}                          &
      \multicolumn{1}{c|}{\ding{52}}                       &
      \multicolumn{1}{c|}{\ding{52}}                       &
      -                                                    &
      -                                                    &
      -                                                    &
      \multicolumn{1}{c|}{-}                               &
      -                                                    &
      17.34                                                &
      43.34                                                &
      31.20                                                &
      \multicolumn{1}{c|}{27.08}                           &
      \multicolumn{1}{c|}{21.30}                           &
      2.90                                                 &
      4.75                                                 &
      1.84                                                 &
      3.64                                                 &
      \multicolumn{1}{c|}{1.02}                            &
      3.79                                                   \\
      &
      \multicolumn{1}{c|}{\ding{52}}                       &
      \multicolumn{1}{c|}{\ding{52}}                       &
      \ding{52}                                            &
      -                                                    &
      -                                                    &
      \multicolumn{1}{c|}{-}                               &
      -                                                    &
      14.38                                                &
      41.55                                                &
      30.90                                                &
      \multicolumn{1}{c|}{25.16}                           &
      \multicolumn{1}{c|}{16.77}                           &
      5.13                                                 &
      19.15                                                &
      0.20                                                 &
      13.81                                                &
      \multicolumn{1}{c|}{0.18}                            &
      -                                                      \\ \midrule
      \multirow{2}{*}{Gemma-2-9B}                          &
      \multicolumn{1}{c|}{\ding{52}}                       &
      \multicolumn{1}{c|}{\ding{52}}                       &
      -                                                    &
      -                                                    &
      -                                                    &
      \multicolumn{1}{c|}{-}                               &
      -                                                    &
      35.39                                                &
      58.24                                                &
      52.15                                                &
      \multicolumn{1}{c|}{43.14}                           &
      \multicolumn{1}{c|}{36.69}                           &
      16.40                                                &
      9.76                                                 &
      2.63                                                 &
      9.02                                                 &
      \multicolumn{1}{c|}{2.57}                            &
      -                                                      \\
      &
      \multicolumn{1}{c|}{\ding{52}}                       &
      \multicolumn{1}{c|}{\ding{52}}                       &
      \ding{52}                                            &
      -                                                    &
      -                                                    &
      \multicolumn{1}{c|}{-}                               &
      -                                                    &
      33.56                                                &
      55.89                                                &
      49.45                                                &
      \multicolumn{1}{c|}{41.48}                           &
      \multicolumn{1}{c|}{35.09}                           &
      14.69                                                &
      27.60                                                &
      0.06                                                 &
      22.68                                                &
      \multicolumn{1}{c|}{0.07}                            &
      -                                                      \\ \midrule
      \multirow{2}{*}{
        \begin{tabular}[c]{@{}c@{}}Llama-3.1-8B- \\ Instruct
      \end{tabular}} &
      \multicolumn{1}{c|}{\ding{52}}                       &
      \multicolumn{1}{c|}{\ding{52}}                       &
      -                                                    &
      -                                                    &
      -                                                    &
      \multicolumn{1}{c|}{-}                               &
      -                                                    &
      33.96                                                &
      56.00                                                &
      48.67                                                &
      \multicolumn{1}{c|}{40.45}                           &
      \multicolumn{1}{c|}{34.56}                           &
      16.78                                                &
      9.31                                                 &
      6.57                                                 &
      8.90                                                 &
      \multicolumn{1}{c|}{6.48}                            &
      -                                                      \\
      &
      \multicolumn{1}{c|}{\ding{52}}                       &
      \multicolumn{1}{c|}{\ding{52}}                       &
      \ding{52}                                            &
      -                                                    &
      -                                                    &
      \multicolumn{1}{c|}{-}                               &
      -                                                    &
      32.25                                                &
      54.21                                                &
      47.05                                                &
      \multicolumn{1}{c|}{39.26}                           &
      \multicolumn{1}{c|}{33.50}                           &
      14.00                                                &
      26.24                                                &
      18.65                                                &
      21.93                                                &
      \multicolumn{1}{c|}{12.62}                           &
      -                                                      \\ \midrule
      \multirow{7}{*}{Qwen2-7B}                            &
      \multicolumn{1}{c|}{\ding{52}}                       &
      \multicolumn{1}{c|}{\ding{52}}                       &
      -                                                    &
      \ding{52}                                            &
      -                                                    &
      \multicolumn{1}{c|}{-}                               &
      -                                                    &
      32.26                                                &
      54.02                                                &
      47.65                                                &
      \multicolumn{1}{c|}{39.44}                           &
      \multicolumn{1}{c|}{33.60}                           &
      15.02                                                &
      20.06                                                &
      4.88                                                 &
      17.99                                                &
      \multicolumn{1}{c|}{4.03}                            &
      -                                                      \\
      &
      \multicolumn{1}{c|}{\ding{52}}                       &
      \multicolumn{1}{c|}{\ding{52}}                       &
      -                                                    &
      -                                                    &
      \ding{52}                                            &
      \multicolumn{1}{c|}{-}                               &
      -                                                    &
      32.23                                                &
      53.26                                                &
      47.40                                                &
      \multicolumn{1}{c|}{39.42}                           &
      \multicolumn{1}{c|}{33.68}                           &
      16.12                                                &
      18.95                                                &
      9.71                                                 &
      16.62                                                &
      \multicolumn{1}{c|}{6.44}                            &
      -                                                      \\
      &
      \multicolumn{1}{c|}{\ding{52}}                       &
      \multicolumn{1}{c|}{\ding{52}}                       &
      -                                                    &
      -                                                    &
      -                                                    &
      \multicolumn{1}{c|}{\ding{52}}                       &
      -                                                    &
      32.71                                                &
      54.94                                                &
      47.48                                                &
      \multicolumn{1}{c|}{40.70}                           &
      \multicolumn{1}{c|}{34.48}                           &
      16.06                                                &
      18.71                                                &
      10.97                                                &
      16.56                                                &
      \multicolumn{1}{c|}{8.17}                            &
      -                                                      \\
      &
      \multicolumn{1}{c|}{\ding{52}}                       &
      \multicolumn{1}{c|}{\ding{52}}                       &
      -                                                    &
      \ding{52}                                            &
      \ding{52}                                            &
      \multicolumn{1}{c|}{-}                               &
      -                                                    &
      31.98                                                &
      53.62                                                &
      47.82                                                &
      \multicolumn{1}{c|}{39.39}                           &
      \multicolumn{1}{c|}{32.27}                           &
      15.75                                                &
      20.95                                                &
      5.95                                                 &
      18.16                                                &
      \multicolumn{1}{c|}{4.13}                            &
      -                                                      \\
      &
      \multicolumn{1}{c|}{\ding{52}}                       &
      \multicolumn{1}{c|}{\ding{52}}                       &
      -                                                    &
      \ding{52}                                            &
      -                                                    &
      \multicolumn{1}{c|}{\ding{52}}                       &
      -                                                    &
      31.89                                                &
      54.39                                                &
      46.46                                                &
      \multicolumn{1}{c|}{39.40}                           &
      \multicolumn{1}{c|}{34.03}                           &
      14.75                                                &
      20.72                                                &
      3.74                                                 &
      17.73                                                &
      \multicolumn{1}{c|}{3.16}                            &
      -                                                      \\
      &
      \multicolumn{1}{c|}{\ding{52}}                       &
      \multicolumn{1}{c|}{\ding{52}}                       &
      -                                                    &
      -                                                    &
      \ding{52}                                            &
      \multicolumn{1}{c|}{\ding{52}}                       &
      -                                                    &
      31.80                                                &
      54.01                                                &
      47.65                                                &
      \multicolumn{1}{c|}{40.11}                           &
      \multicolumn{1}{c|}{34.06}                           &
      16.04                                                &
      19.29                                                &
      6.14                                                 &
      16.78                                                &
      \multicolumn{1}{c|}{4.90}                            &
      -                                                      \\
      &
      \multicolumn{1}{c|}{\ding{52}}                       &
      \multicolumn{1}{c|}{\ding{52}}                       &
      -                                                    &
      \ding{52}                                            &
      \ding{52}                                            &
      \multicolumn{1}{c|}{\ding{52}}                       &
      -                                                    &
      31.77                                                &
      52.86                                                &
      47.39                                                &
      \multicolumn{1}{c|}{38.68}                           &
      \multicolumn{1}{c|}{33.66}                           &
      15.79                                                &
      20.83                                                &
      9.50                                                 &
      18.03                                                &
      \multicolumn{1}{c|}{6.77}                            &
      -                                                      \\ \midrule
      \multirow{4}{*}{Qwen2-7B}                            &
      \multicolumn{1}{c|}{-}                               &
      \multicolumn{1}{c|}{-}                               &
      -                                                    &
      -                                                    &
      -                                                    &
      \multicolumn{1}{c|}{-}                               &
      \ding{52}                                            &
      7.50                                                 &
      8.56                                                 &
      8.43                                                 &
      \multicolumn{1}{c|}{6.58}                            &
      \multicolumn{1}{c|}{4.50}                            &
      10.51                                                &
      10.66                                                &
      22.17                                                &
      9.46                                                 &
      \multicolumn{1}{c|}{16.57}                           &
      25.96                                                  \\
      &
      \multicolumn{1}{c|}{\ding{52}}                       &
      \multicolumn{1}{c|}{\ding{52}}                       &
      -                                                    &
      -                                                    &
      -                                                    &
      \multicolumn{1}{c|}{-}                               &
      \ding{52}                                            &
      33.32                                                &
      55.23                                                &
      49.30                                                &
      \multicolumn{1}{c|}{41.29}                           &
      \multicolumn{1}{c|}{34.69}                           &
      17.78                                                &
      10.04                                                &
      20.11                                                &
      9.13                                                 &
      \multicolumn{1}{c|}{11.13}                           &
      45.13                                                  \\
      &
      \multicolumn{1}{c|}{-}                               &
      \multicolumn{1}{c|}{-}                               &
      \ding{52}                                            &
      -                                                    &
      -                                                    &
      \multicolumn{1}{c|}{-}                               &
      \ding{52}                                            &
      0.02                                                 &
      10.62                                                &
      10.66                                                &
      \multicolumn{1}{c|}{6.93}                            &
      \multicolumn{1}{c|}{4.85}                            &
      12.72                                                &
      25.73                                                &
      1.70                                                 &
      21.49                                                &
      \multicolumn{1}{c|}{1.76}                            &
      37.10                                                  \\
      &
      \multicolumn{1}{c|}{\ding{52}}                       &
      \multicolumn{1}{c|}{\ding{52}}                       &
      \ding{52}                                            &
      -                                                    &
      -                                                    &
      \multicolumn{1}{c|}{-}                               &
      \ding{52}                                            &
      31.31                                                &
      51.45                                                &
      48.57                                                &
      \multicolumn{1}{c|}{39.05}                           &
      \multicolumn{1}{c|}{33.69}                           &
      13.29                                                &
      26.35                                                &
      6.17                                                 &
      21.95                                                &
      \multicolumn{1}{c|}{5.56}                            &
      42.66                                                  \\ \bottomrule
    \end{tabular}%
  }
  \caption{Comprehensive BLEU scores for machine translation experiments.}
  \label{tab:all_bleu_score}
\end{sidewaystable*}

%% file: tables/low-resource_all.tex
\begin{table*}[htb!]
  \centering
  \small
  \begin{subtable}{\linewidth}
    \input{tables/low-resource_mt}
  \end{subtable}
  \vspace{1mm}

  \begin{subtable}[t]{0.4\linewidth}
    \input{tables/low-resource_ner}
  \end{subtable}
  \hspace{10mm}
  \begin{subtable}[t]{0.4\linewidth}
    \input{tables/low-resource_punc}
  \end{subtable}

  \caption{Ablation study results showing model performance across varying ratios of Hanja (Hj) to Classical Chinese (Lzh) training data for (a) machine translation measured by BLEU score, (b) named entity recognition measured by F1 score, and (c) punctuation restoration measured by F1 score. Ratios range from using only Lzh data (0:1) to the full Hj:Lzh ratio for each task. $\dagger$ denotes evaluation on augmented data.}
  \label{tab:low-resource_all}
\end{table*}

%% file: tables/low-resource_mt.tex
\resizebox{\linewidth}{!}{%
  \begin{tabular}{@{}lcccccccccc@{}}
    \toprule
    \multirow{3}{*}{\textbf{
        \begin{tabular}[l]{@{}l@{}}Train Data \\ Ratio\\ (Hj : Lzh)
    \end{tabular}}} &
    \multicolumn{4}{c}{\textbf{Hj$^\text{R}$}}                  &
    \textbf{Hj$^\text{L}$}                                      &
    \multicolumn{5}{c}{\textbf{Lzh}}                                                                                                            \\ \cmidrule(l){2-11}
    &
    \multicolumn{2}{c}{\textbf{AJD}}                            &
    \textbf{DRS}                                                &
    \textbf{DRRI}                                               &
    \textbf{KLC}                                                &
    \textbf{OCDB}                                               &
    \multicolumn{2}{c}{\textbf{NiuTrans}}                       &
    \multicolumn{2}{c}{\textbf{WYWMT}}                                                                                                          \\ \cmidrule(l){2-11}
    &
    \textbf{Hj-En}                                              &
    \textbf{Hj-Ko}                                              &
    \textbf{Hj-Ko}                                              &
    \textbf{Hj-Ko}                                              &
    \textbf{Hj-Ko}                                              &
    \textbf{Lzh-Ko}                                             &
    \textbf{Lzh-Ko}                                             &
    \textbf{Lzh-Zh}                                             &
    \textbf{Lzh-Ko}                                             &
    \textbf{Lzh-Zh}                                                                                                                             \\ \midrule
    $0.496 : 0$                                                 & 33.15 & 55.30 & 48.65 & 40.65 & 33.07 & 16.13 & 9.42  & 15.13 & 7.33  & 8.74  \\
    $0.496 : 1$                                                 & 31.52 & 52.83 & 47.04 & 39.33 & 33.91 & 14.26 & 26.06 & 1.21  & 21.68 & 0.86  \\ \midrule
    $2^{-2} : 0$                                                & 31.26 & 52.01 & 47.15 & 39.21 & 31.80 & 15.72 & 9.93  & 20.47 & 8.45  & 11.81 \\
    $2^{-2} : 1$                                                & 29.32 & 51.29 & 45.37 & 37.54 & 32.28 & 14.18 & 25.69 & 8.30  & 22.09 & 7.53  \\ \midrule
    $2^{-3} : 0$                                                & 29.00 & 51.01 & 45.42 & 36.02 & 29.15 & 14.68 & 9.15  & 19.75 & 7.55  & 11.73 \\
    $2^{-3} : 1$                                                & 26.95 & 48.38 & 42.75 & 36.83 & 30.62 & 12.94 & 26.13 & 10.78 & 21.66 & 10.09 \\ \midrule
    $2^{-4} : 0$                                                & 26.63 & 47.25 & 39.72 & 33.36 & 25.35 & 12.91 & 8.42  & 22.64 & 7.06  & 14.67 \\
    $2^{-4} : 1$                                                & 24.18 & 47.51 & 37.13 & 34.01 & 28.96 & 13.71 & 25.92 & 8.38  & 22.20 & 9.05  \\ \midrule
    $2^{-5} : 0$                                                & 23.20 & 43.70 & 37.25 & 30.97 & 23.76 & 11.52 & 8.35  & 26.19 & 7.28  & 18.17 \\
    $2^{-5} : 1$                                                & 20.76 & 44.76 & 35.37 & 29.93 & 27.94 & 13.28 & 26.05 & 4.10  & 21.88 & 4.46  \\ \midrule
    $0 : 0$                                                     & -     & -     & -     & -     & -     & -     & -     & -     & -     & -     \\
    $0 : 1$                                                     & 0.02  & 10.96 & 10.35 & 7.22  & 4.85  & 12.93 & 26.25 & 5.75  & 21.60 & 6.18  \\ \bottomrule
  \end{tabular}
}
\caption{MT (BLEU)}
\label{tab:low-resource_mt}

%% file: tables/low-resource_ner.tex
\resizebox{\columnwidth}{!}{%
  \begin{tabular}{@{}lccc@{}}
    \toprule
    \multirow{2}{*}{\textbf{
        \begin{tabular}[c]{@{}l@{}}Train Data Ratio \\ (Hj : Lzh)
    \end{tabular}}} & \textbf{Hj$^{\text{R}}$} & \textbf{Hj$^{\text{L}}$} & \textbf{Lzh}                                             \\ \cmidrule(l){2-4}
    & \textbf{AJD}             & \textbf{KLC}             & \textbf{GLNER} \\ \midrule
    $20.5 : 0$                                                & 97.53                    & 83.55                    & 66.15          \\
    $20.5 : 1$                                                & 97.45                    & 84.22                    & 87.68          \\ \midrule
    $2^4 : 0$                                                 & 97.39                    & 83.42                    & 65.92          \\
    $2^4 : 1$                                                 & 97.40                    & 83.71                    & 87.83          \\ \midrule
    $2^3 : 0$                                                 & 97.14                    & 82.41                    & 65.82          \\
    $2^3 : 1$                                                 & 97.00                    & 82.39                    & 87.77          \\ \midrule
    $2^2 : 0$                                                 & 96.63                    & 80.94                    & 65.28          \\
    $2^2 : 1$                                                 & 96.53                    & 80.43                    & 87.54          \\ \midrule
    $2^1 : 0$                                                 & 96.07                    & 78.70                    & 64.83          \\
    $2^1 : 1$                                                 & 95.81                    & 78.30                    & 87.20          \\ \midrule
    $1 : 0$                                                   & 95.33                    & 76.25                    & 64.03          \\
    $1 : 1$                                                   & 94.81                    & 77.19                    & 87.06          \\ \midrule
    $2^{-1} : 0$                                              & 94.26                    & 72.48                    & 62.37          \\
    $2^{-1} : 1$                                              & 93.74                    & 74.16                    & 86.83          \\ \midrule
    $2^{-2} : 0$                                              & 92.94                    & 68.82                    & 60.48          \\
    $2^{-2} : 1$                                              & 92.35                    & 72.46                    & 86.83          \\ \midrule
    $2^{-3} : 0$                                              & 90.44                    & 65.54                    & 56.76          \\
    $2^{-3} : 1$                                              & 90.26                    & 69.15                    & 86.58          \\ \midrule
    $2^{-4} : 0$                                              & 85.64                    & 62.31                    & 52.14          \\
    $2^{-4} : 1$                                              & 87.58                    & 73.10                    & 86.69          \\ \midrule
    $2^{-5} : 0$                                              & 73.97                    & 41.18                    & 34.32          \\
    $2^{-5} : 1$                                              & 85.99                    & 73.31                    & 86.60          \\ \midrule
    $0 : 0$                                                   & -                        & -                        & -              \\
    $0 : 1$                                                   & 81.32                    & 72.61                    & 86.48          \\ \bottomrule
  \end{tabular}
}
\caption{NER (F1)}
\label{tab:low-resource_ner}

%% file: tables/low-resource_punc.tex
\resizebox{\columnwidth}{!}{%
  \begin{tabular}{@{}lccc@{}}
    \toprule
    \multirow{2}{*}{\textbf{
        \begin{tabular}[c]{@{}l@{}}Train Data Ratio \\ (Hj : Lzh)
    \end{tabular}}} & \textbf{Hj$^{\text{R}}$} & \textbf{Hj$^{\text{L}}$} & \textbf{Lzh}                                             \\ \cmidrule(l){2-4}
    & \textbf{AJD}             & \textbf{KLC}             & \textbf{WYWEB} \\ \midrule
    $4.36 : 0$                                                & 88.61                    & 87.76                    & 78.02          \\
    $4.36 : 1$                                                & 88.57                    & 87.91                    & 85.28          \\ \midrule
    $2^2 : 0$                                                 & 88.54                    & 87.74                    & 78.12          \\
    $2^2 : 1$                                                 & 88.54                    & 87.85                    & 85.42          \\ \midrule
    $2^1 : 0$                                                 & 87.99                    & 87.17                    & 77.89          \\
    $2^1 : 1$                                                 & 87.96                    & 87.27                    & 85.76          \\ \midrule
    $1 : 0$                                                   & 87.39                    & 86.65                    & 77.62          \\
    $1 : 1$                                                   & 87.25                    & 86.77                    & 85.76          \\ \midrule
    $2^{-1} : 0$                                              & 86.65                    & 86.00                    & 77.35          \\
    $2^{-1} : 1$                                              & 86.67                    & 86.36                    & 85.84          \\ \midrule
    $2^{-2} : 0$                                              & 85.95                    & 85.28                    & 76.95          \\
    $2^{-2} : 1$                                              & 85.90                    & 85.85                    & 85.88          \\ \midrule
    $2^{-3} : 0$                                              & 84.93                    & 84.19                    & 76.31          \\
    $2^{-3} : 1$                                              & 85.10                    & 85.26                    & 85.93          \\ \midrule
    $2^{-4} : 0$                                              & 83.60                    & 82.20                    & 74.87          \\
    $2^{-4} : 1$                                              & 83.67                    & 84.29                    & 85.92          \\ \midrule
    $2^{-5} : 0$                                              & 81.16                    & 79.17                    & 72.89          \\
    $2^{-5} : 1$                                              & 81.35                    & 83.45                    & 85.87          \\ \midrule
    $0 : 0$                                                   & -                        & -                        & -              \\
    $0 : 1$                                                   & 78.36                    & 80.66                    & 85.83          \\ \bottomrule
  \end{tabular}
}
\caption{PR (F1)}
\label{tab:low-resource_punc}

%% file: figures/kanbun-low.tex
\begin{figure}[t!]
  \centering
  \includegraphics[width=\columnwidth]{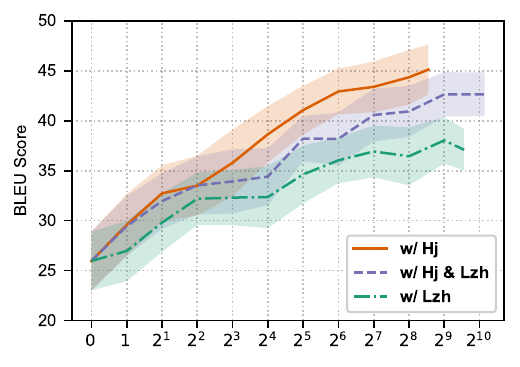}
  \caption{Performance comparison of Kanbun-Korean translation models with varying amounts of additional training data. The $x$-axis shows the ratio of additional data to Kanbun data in $\log_2$ scale, and the $y$-axis shows BLEU scores with 95\% confidence intervals indicated by shaded regions.}
  \label{fig:kanbun-low}
\end{figure}

%% file: figures/vocab-heatmap-all.tex
\begin{figure*}[htb!]
  \centering
  \begin{subfigure}{0.45\linewidth}
    \includegraphics[width=\linewidth]{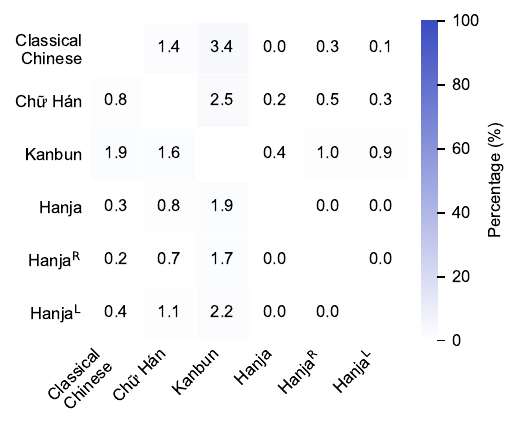}
    \caption{100\%}
    \label{fig:subfig1}
  \end{subfigure}
  \hspace{10mm}
  \begin{subfigure}{0.45\linewidth}
    \includegraphics[width=\linewidth]{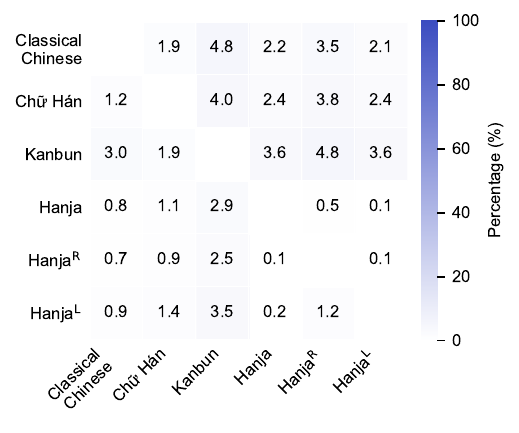}
    \caption{99.9\%}
    \label{fig:subfig2}
  \end{subfigure}
  \vspace{1mm}
  \begin{subfigure}{0.45\linewidth}
    \includegraphics[width=\linewidth]{images/vocab-heatmap-99.00.pdf}
    \caption{99\%}
    \label{fig:subfig3}
  \end{subfigure}
  \hspace{10mm}
  \begin{subfigure}{0.45\linewidth}
    \includegraphics[width=\linewidth]{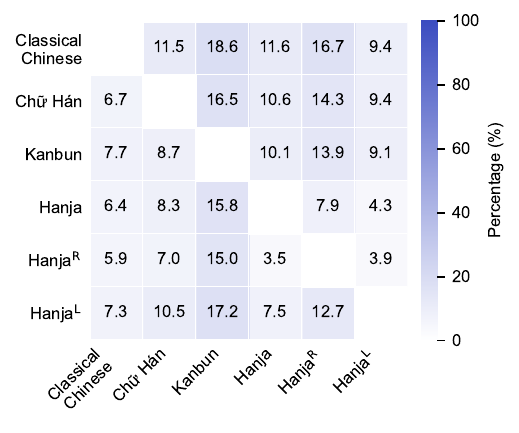}
    \caption{95\%}
    \label{fig:subfig4}
  \end{subfigure}
  \caption{Character divergence patterns across writing systems at different frequency thresholds.}
  \label{fig:vocab-heatmap-all}
\end{figure*}

%% file: figures/low_resource_loss.tex
\begin{figure*}[htb!]
  \centering
  \begin{subfigure}{0.45\linewidth}
    \centering
    \includegraphics[width=\linewidth]{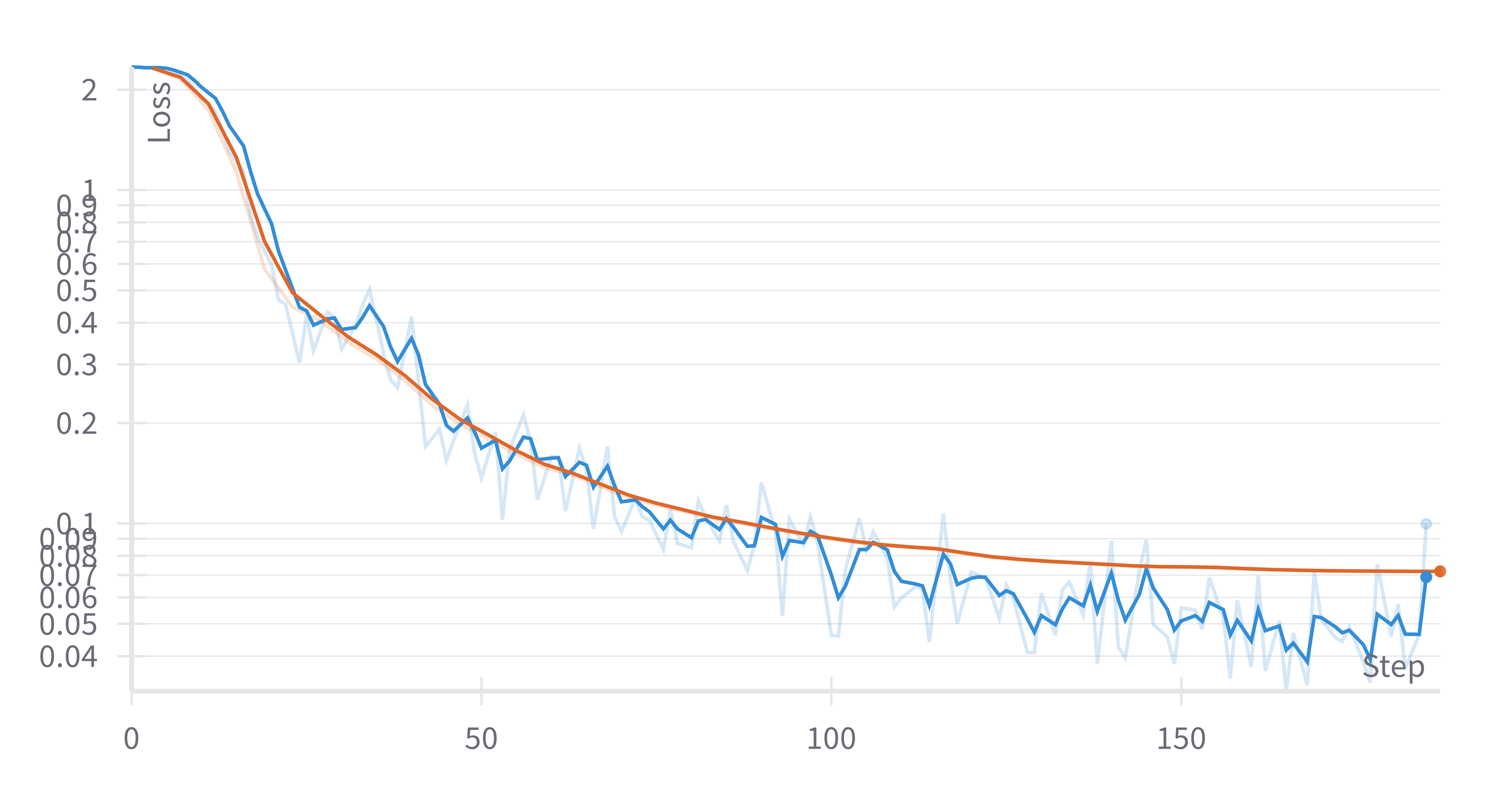}
    \caption{NER (1/16 of original Hanja data)}
    \label{fig:ner_1_16}
  \end{subfigure}
  \hspace{10mm}
  \begin{subfigure}{0.45\linewidth}
    \centering
    \includegraphics[width=\linewidth]{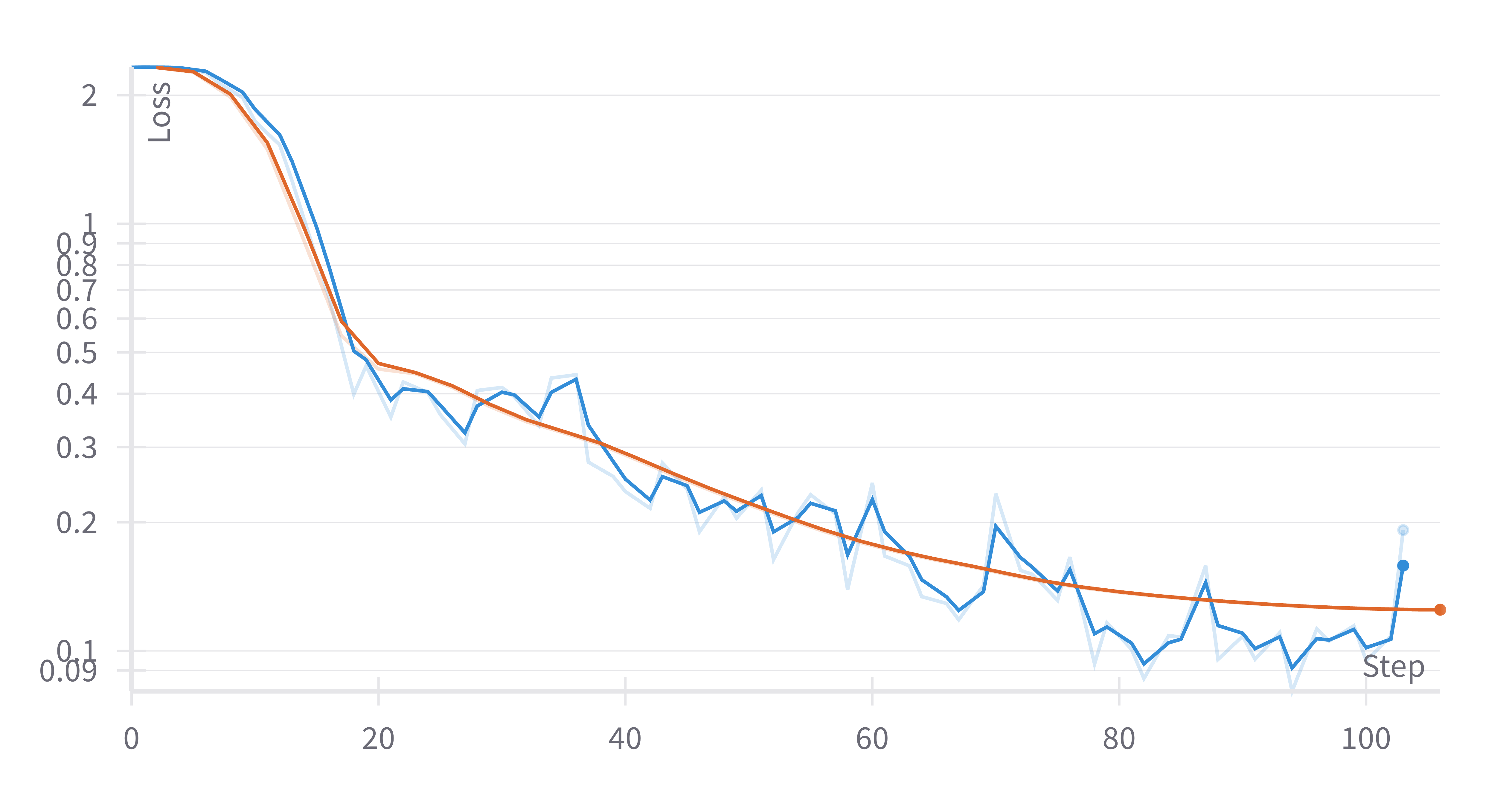}
    \caption{NER (1/32 of original Hanja data)}
    \label{fig:ner_1_32}
  \end{subfigure}
  \vspace{1mm}
  \begin{subfigure}{0.45\linewidth}
    \centering
    \includegraphics[width=\linewidth]{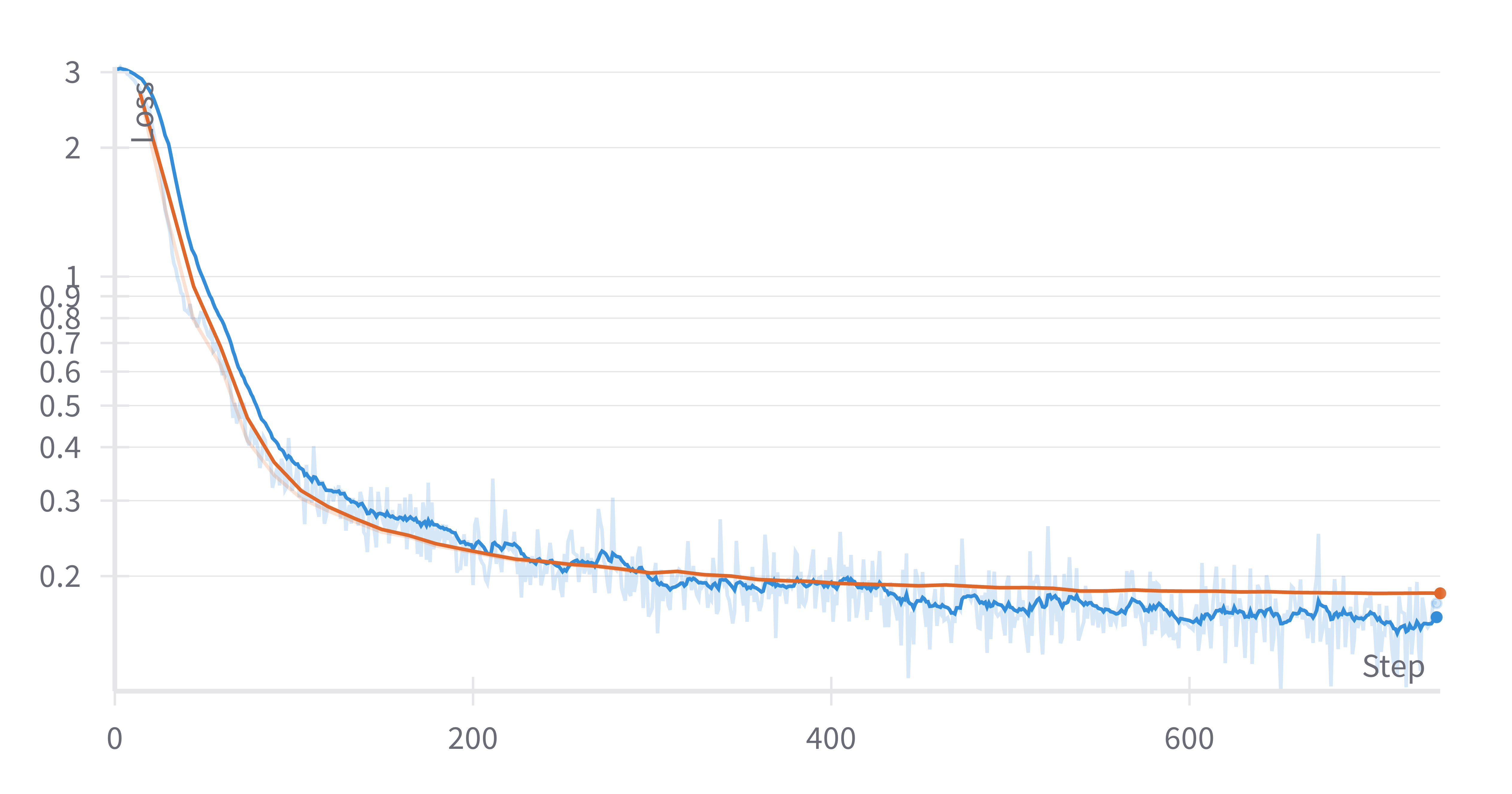}
    \caption{PR (1/16 of original Hanja data)}
    \label{fig:pr_1_16}
  \end{subfigure}
  \hspace{10mm}
  \begin{subfigure}{0.45\linewidth}
    \centering
    \includegraphics[width=\linewidth]{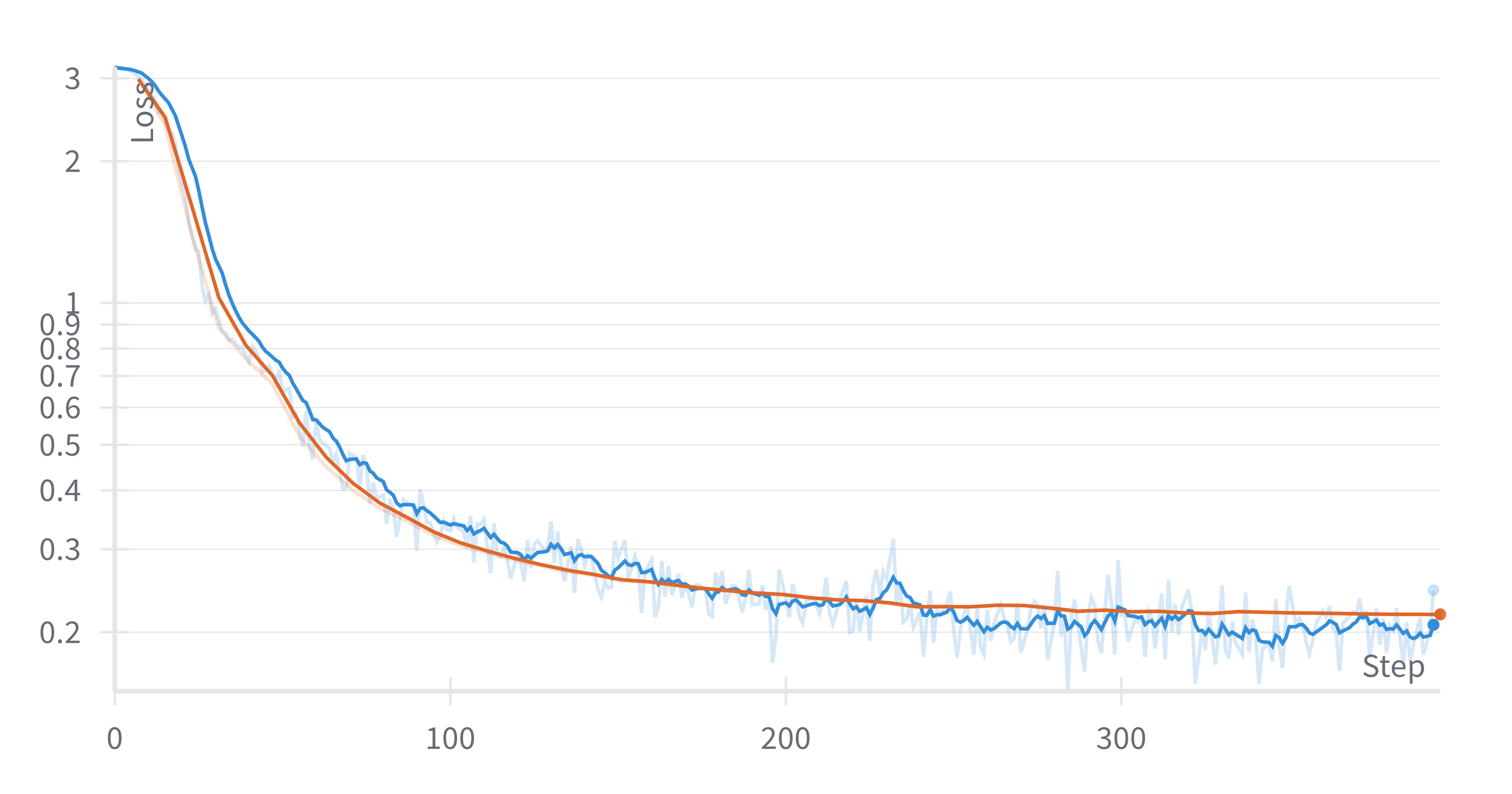}
    \caption{PR (1/32 of original Hanja data)}
    \label{fig:pr_1_32}
  \end{subfigure}
  \caption{Evaluation loss curves for NER and PR tasks in low-resource settings. Blue lines and orange lines represent training loss and validation loss, respectively.}
  \label{fig:low_resource_loss}
\end{figure*}

%% file: tables/error_analysis.tex
\begin{sidewaystable*}
  \small
  \centering
  \rowcolors{2}{gray!10}{white}
  \renewcommand{\arraystretch}{1.3}
  \begin{tabular}{@{}p{2.5cm}p{4.5cm}p{4.5cm}p{4.5cm}p{4.5cm}@{}}
    \toprule
    \textbf{Error Type}                                                                                                                                         &
    \textbf{Source Hanja Text}                                                                                                                                  &
    \textbf{Ground-truth Korean Translation}                                                                                                                    &
    \textbf{Prediction (Model w/o Lzh)}                                                                                                                         &
    \textbf{Prediction (Model w/ Lzh)}                                                                                                                            \\ \midrule
    Modernization of classical terms                                                                                                                            &
    諭黃海道觀察使曰: ``延安付處瓔, 依住家舍, 令所在邑隨宜修理, 當身及奴婢幷五口, 月給糧料、酒肉、饌具。''                                                                                                  &
    황해도 관찰사(黃海道觀察使)에게 유시(諭示)하기를, ``연안(延安)에 부처(付處)한 이영(李瓔)이 의지해 살 가옥을 소재한 고을로 하여금 적절히 수리하게 하고, 그 자신과 노비(奴婢) 5구(口)에게 아울러 매월 양료(糧料)·주육(酒肉)·찬구(饌具)를 지급하라.'' 하였다.    &
    황해도 관찰사(黃海道觀察使)에게 유시(諭示)하기를, ``연안(延安)에 부처(付處)한 이영(李瓔)은 그의 거처하는 집을 소재읍(所在邑)에서 적당히 수리하게 하고, 그 자신과 노비(奴婢) 아울러 5구(口)에게는 달마다 양료(糧料)와 술·고기·찬구(饌具)를 주도록 하라.'' 하였다. &
    황해도 관찰사에게 유시하기를, ``연안(延安)에 부처(付處)한 이영(李瓔)은 그가 사는 집에 의거하여 거처하게 하고, 소재(所在)하는 고을로 하여금 적당히 수리하게 하며, 당신과 노비와 아울러 5명에게는 매월 양식·술·고기·반찬을 주도록 하라.'' 하였다.               \\
    Loss of Korea-specific concepts                                                                                                                             &
    政院啓曰: ``經理都監郞廳來言: '自中原, 表裏出來于經理處云, 故經理明日欲迎於郊外, 黃儀仗取來' 云。''                                                                                                  &
    정원이 아뢰었다. ``경리 도감 낭청이 와서 `중국에서 표리(表裏)를 경리에게 보내왔다 하므로 경리가 내일 교외에서 맞이하려고 하는데 황의장(黃儀仗)을 가져 오라고 했다.' 하였습니다.''                                                   &
    정원이 아뢰었다. ``경리 도감 낭청이 와서 말하기를 `중원에서 표리(表裏)가 경리에게 나왔다고 하므로 경리가 내일 교외에서 맞이하려고 하니 황의장(黃儀仗)을 가져오라.'고 하였습니다.''                                                   &
    정원이 아뢰었다. ``경리 도감 낭청이 와서 말하기를 `중원에서 표리(表裏)가 나와 경리처에 왔다.'고 하므로 경리가 내일 교외에서 맞이하고자 하니, 의장(儀仗)을 가져오라고 하였다.''                                                      \\
    Name translation errors                                                                                                                                     &
    臺諫啓前事, 命遞崔連孫、尹滂、成希仲、閔慶安, 餘不允。                                                                                                                               &
    대간이 전의 일을 아뢰니 최연손·윤방·성희중·민경안은 체직하도록 하고, 나머지는 윤허하지 않았다.                                                                                                      &
    대간이 전의 일을 아뢰니, 최연손·윤방·성희중·민경안은 체직시키고 나머지는 윤허하지 않았다.                                                                                                         &
    대간이 전의 일을 아뢰니, 최연손·윤팽·성희중·민경안은 체직시키라 명하고 나머지는 윤허하지 않았다.                                                                                                       \\ \bottomrule
  \end{tabular}
  \caption{Examples of translation errors when incorporating Classical Chinese resources}
  \label{tab:error_analysis}
\end{sidewaystable*}